\documentclass[10pt,twocolumn,letterpaper]{article}
\usepackage{iccv}              % To produce the CAMERA-READY version

\usepackage[linesnumbered,lined,boxed,commentsnumbered,ruled,vlined,]{algorithm2e}
\usepackage{mathtools}
\usepackage{multirow}
\usepackage{colortbl}
\usepackage{pifont}
\usepackage{multirow}
\usepackage{graphicx}
\usepackage[normalem]{ulem}
\useunder{\uline}{\ul}{}
\definecolor{iccvblue}{rgb}{0.21,0.49,0.74}
\usepackage[pagebackref,breaklinks,colorlinks,allcolors=iccvblue]{hyperref}

\def\paperID{13783} % *** Enter the Paper ID here
\def\confName{ICCV}
\def\confYear{2025}

\title{SFOOD: A Multimodal Benchmark for Comprehensive Food Attribute Analysis Beyond RGB with Spectral Insights}

\author{
  Zhenbo Xu$^{1,2,*}$,
  Jinghan Yang$^{1}$,
  Gong Huang$^{2}$,
  Jiqing Feng$^{2}$, 
  Liu Liu$^{2}$, \\
  Ruihan Sun$^{2}$,
  Ajin Meng$^{2}$,
  Zhuo Zhang$^{3}$,
  Zhaofeng He$^{1}$\\
  $^{1}$Beijing University of Posts and Telecommunications \\
  $^{2}$ShiFang Technology Inc., Hangzhou, China \\
  $^{3}$Specrizion Technology Co.Ltd, Wuxi, China
}

\newcommand{\cmark}{\textcolor{green}{\ding{51}}} % Check mark
\newcommand{\xmark}{\textcolor{red}{\ding{55}}}   % Cross mark

\begin{document}

\twocolumn[{%
\renewcommand\twocolumn[1][]{#1}%
\maketitle
\begin{center}
    \centering
    \captionsetup{type=figure}
    \includegraphics[width=1.0\textwidth]{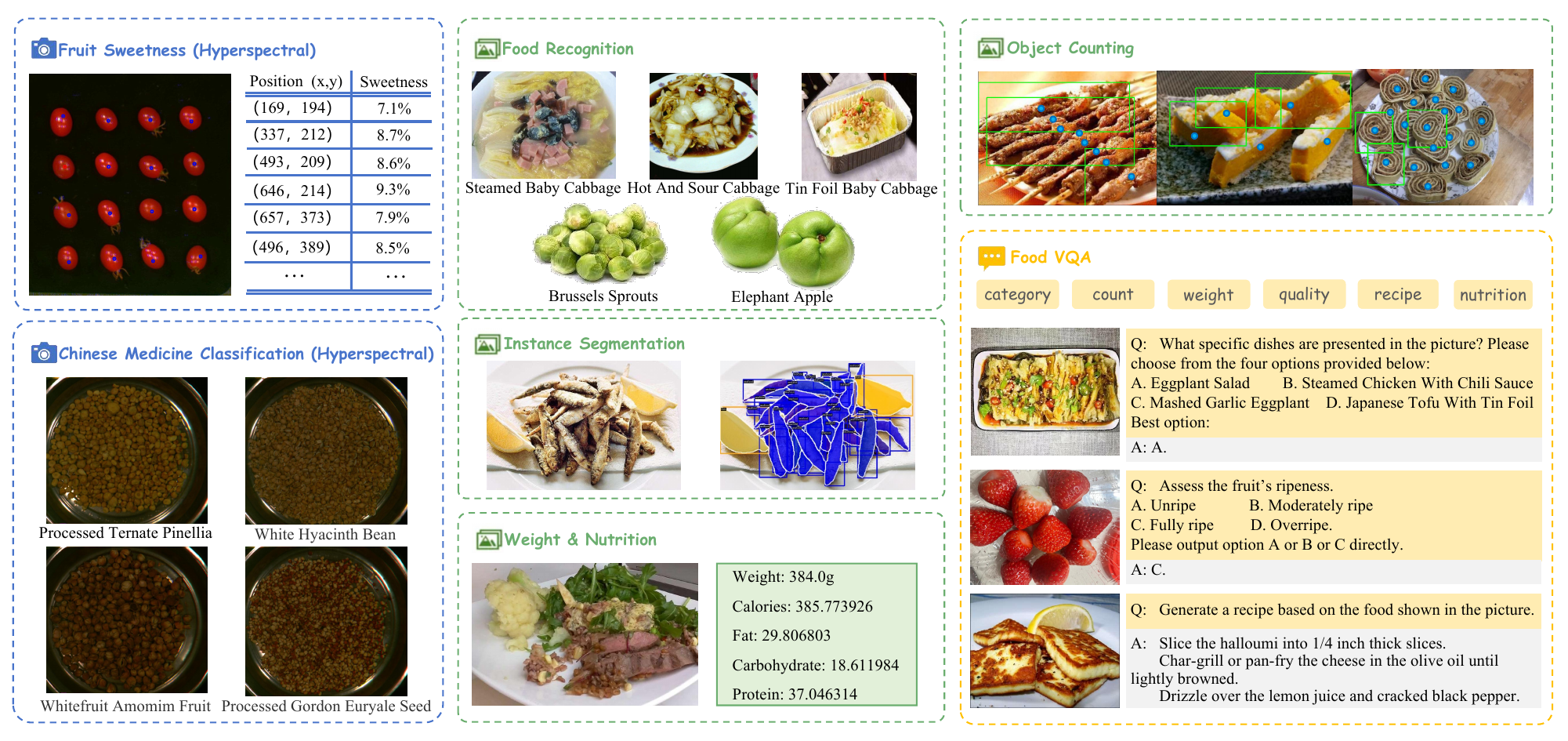}
    \captionof{figure}{SFOOD contains 6 different food analysis tasks, 17 main categories, 3,266 food subcategories, and 2.35 million data.}
\end{center}%
}]

\maketitle
\begin{abstract}
With the rise and development of computer vision and LLMs, intelligence is everywhere, especially for people and cars. However, for tremendous food attributes (such as origin, quantity, weight, quality, sweetness, etc.), existing research still mainly focuses on the study of categories. The reason is the lack of a large and comprehensive benchmark for food. Besides, many food attributes (such as sweetness, weight, and fine-grained categories) are challenging to accurately percept solely through RGB cameras. To fulfill this gap and promote the development of intelligent food analysis, in this paper, we built the first large-scale spectral food (SFOOD) benchmark suite. We spent a lot of manpower and equipment costs to organize existing food datasets and collect hyperspectral images of hundreds of foods, and we used instruments to experimentally determine food attributes such as sweetness and weight. The resulting benchmark consists of 3,266 food categories and 2,351 k data points for 17 main food categories. Extensive evaluations find that: (i) Large-scale models are still poor at digitizing food. Compared to people and cars, food has gradually become one of the most difficult objects to study; (ii) Spectrum data are crucial for analyzing food properties (such as sweetness). Our benchmark will be open source and continuously iterated for different food analysis tasks.
\end{abstract}    
\begin{table*}[]
\centering
\resizebox{1.0\textwidth}{!}{%
\begin{tabular}{l|cccccccc|cc}
\hline
Current Datasets                                                                                                                                                                                                                         & Recognition & Detection & Counting & \begin{tabular}[c]{@{}c@{}}Weight \&\\ Nutrition\end{tabular} & Recipe & Quality & \begin{tabular}[c]{@{}c@{}}QA \&\\ Captioning\end{tabular} & \begin{tabular}[c]{@{}c@{}}Spectral \\ Data\end{tabular} & \begin{tabular}[c]{@{}c@{}}Max Class\\ Count\end{tabular} & \begin{tabular}[c]{@{}c@{}}Max Data\\ Count\end{tabular} \\ \hline
\begin{tabular}[c]{@{}l@{}}Food-101 \cite{bossard2014food}, ChineseFoodNet \cite{hou2017vegfru}, ISIA Food-500 \cite{min2020isia}, \\ FoodX-251 \cite{kaur2019foodx}, Food2K \cite{min2023large}, VegFru \cite{hou2017vegfru}, \\ UNICT-FD1200 \cite{farinella2016retrieval}, Vireo-Food 172 \cite{chen2016deep},  Food975 \cite{zhou2016fine},\\ Fruits-262 \cite{minut2021fruits}, GroceryStore \cite{klasson2019hierarchical}, Products10k \cite{bai2020products},\\ PlantVillage \cite{mohanty2016using}, CWD30 \cite{ilyas2023cwd30}, CNH-98 \cite{xu2021multiple}\end{tabular} & \cmark           & \xmark         & \xmark        & \xmark                                                             & \xmark  & \xmark      & \xmark       & \xmark                                                        & 2,000                                                     & 1,036k                                                   \\ \hline
\begin{tabular}[c]{@{}l@{}}UEC Food-256 \cite{kawano2015automatic}, UEC-FoodPix \cite{okamoto2021uec}, FoodSeg103 \cite{wu2021large}, \\ CropAndWeed \cite{steininger2023cropandweed}, RiceSeedling \cite{tseng2022rice}, PlantDoc \cite{singh2020plantdoc}\end{tabular}                                                                                                                             & \xmark           & \cmark         & \xmark        & \xmark                                                             & \xmark      & \xmark       & \xmark         & \xmark                                                & 256                                                       & 32k                                                      \\ \hline
FSC-147 \cite{ranjan2021learning}, RiceSeedlingCount \cite{wu2019automatic}, OmniCount (*)  \cite{mondal2024omnicount}                                                                                                                        & \xmark           & \xmark         & \cmark        & \xmark                                                             & \xmark      & \xmark       & \cmark       & \xmark                                                 & 191                                                       & 30k                                                       \\ \hline
FooDD \cite{pouladzadeh2015foodd}, Nutrition5K \cite{thames2021nutrition5k}                                                                                                                                                                                                                      & \xmark           & \xmark         & \xmark        & \cmark                                                             & \xmark      & \xmark       & \xmark      & \xmark                                                   & 250                                                       & 5k                                                       \\ \hline
Pic2kcal \cite{ruede2021multi}, Recipe1M+ \cite{marin2021recipe1m+}, RecipeNLG \cite{bien2020recipenlg}                                                                                                                                                                                                                    & \xmark           & \xmark         & \xmark        & \xmark                                                             & \cmark      & \xmark       & \cmark     & \xmark                                                    & Undefined                                                 & 2,231k                                                   \\ \hline
Rice-5 \cite{izquierdo2020visible}, FruitNet \cite{meshram2022fruitnet}, FruitQ \cite{abayomi2024fruitq}                                                                                                                                                                                                                 & \xmark           & \xmark         & \xmark        & \xmark                                                            & \xmark      & \cmark       & \xmark      & \xmark                                                   & 11                                                        & 27k                                                      \\ \hline
HSIFoodIngr-64 \cite{xia2023hsifoodingr}, Honey-32 \cite{noviyanto2017honey}                                                                                                                                                                                                                & \xmark           & \xmark         & \xmark       & \xmark                                                             & \xmark      & \xmark       & \xmark & \cmark                                                        & 64                                                        & 3k                                                       \\ \hline
Food-Sky \cite{zhou2024foodsky} (*)                                                                                                                                                                                                     & \xmark           & \xmark         & \xmark       & \xmark                                                             & \xmark      & \xmark       & \cmark & \xmark                                                        & Undefined                                                        & 811k                                                       \\ \hline
Uni-Food \cite{xia2023hsifoodingr} (*)                                                                                                                                                                                                            &\cmark           & \xmark         & \xmark       & \cmark                                                             & \cmark      & \xmark       & \cmark & \xmark                                                      & Undefined                                                        & 100k                                                       \\ \hline
SFOOD (ours)                                                                                                                                                                                                                             & \cmark           & \cmark         & \cmark        & \cmark                                                             & \cmark      & \cmark & \cmark      & \cmark                                                        & \textbf{3,266}                                                   & \textbf{2,351k}                                                   \\ \hline
\end{tabular}
}
\caption{Comparison with current food datasets. Our SFOOD benchmark is more large-scale and comprehensive than current datasets. * means not open-source yet.}
\label{dataset_comparison_table}
\end{table*}

\section{Introduction}
\label{sec:intro}

The relationship between food and people is complex, affecting health and life expectancy \cite{Adler2023}. With the rapid development of AI technology and the growing concern for food safety and dietary health, the multi-dimensional attributes of food are expected to be digitally decomposed, creating the possibility of deciphering the connection between them. However, based solely on traditional RGB cameras, even the state-of-the-art large vision language models (LVLMs) and humans have been unable to achieve satisfactory performance in the multi-dimensional attributes and fine-grained analysis of food. Extensive research on food analysis beyond the visible spectrum is crucial for food digitization and understanding the impact of food on human development.

Food has many different attributes, including type, appearance, sweetness, origin, weight, nutritional content, and more. The complexity of food makes it a challenging subject compared to other objects \cite{wu2021large}. Existing open-source food datasets \cite{bossard2014food,chen2016deep,min2023large} have significant limitations in terms of construction dimensions, coverage, and data size, as shown in Table \ref{dataset_comparison_table}. Specifically, these limitations manifest in three main deficiencies: (a) Focusing on basic visual tasks familiar to people, such as classification and detection, but lack the task of analysing attributes beyond the scope of human vision, such as sweetness. (b) Most datasets concentrate on a single food-related task and have not established comprehensive benchmarks that can assess multi-dimensional attributes. (c) Data collection involving dimensions like sweetness, weight, and spectra often requires specific precision instruments and measurements by professionals, significantly increasing the cost of data collection and annotation. These factors collectively result in the absence of a benchmark that can comprehensively assess the multi-dimensional attributes of food up until now.

\begin{figure}[!t]
\centering
\includegraphics[width=1.0\linewidth]{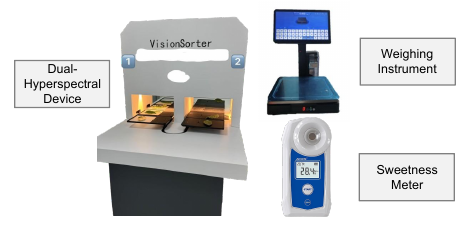}
\caption{Our currently adopted devices to build SFOOD. From left to right: the data collection device with dual hyperspectral cameras, the weighing instruments, the sweetness meters. }
\label{food_capture_device}
% \vspace{-4mm}
\end{figure}

To enhance our understanding of the multi-dimensional attributes of food, we construct and open-source a comprehensive multi-modal benchmark suite that encompasses a wide range of food attributes, standardizing the data categories and several food-related tasks. Initially, we classify common food products into 17 distinct parent categories by following previous datasets and incorporating the principal food groups defined by the United States Department of Agriculture (USDA). These categories are: bakery, cake \& pie, egg, seafood, meat, noodles \& steamed pasta, grain \& cereal, soup, vegetable, fruit, bean product, dairy product, dessert \& beverage, sauce \& dressing, packaging food, mixed food, and Chinese herbal medicine. In addition, we have employed a variety of diverse food acquisition devices to gather data across various food attributes, as depicted in Fig. \ref{food_capture_device}, including data collection platform with hyperspectral cameras, precision weighing instruments for recording exact weight measurements, and sweetness meters for assessing the Brix value of food. The resulting benchmark comprises six primary tasks: fine-grained image classification, instance segmentation, counting, visual question answering, spectral food sweetness analysis, and spectral Chinese herbal medicine classification. Our benchmark suite is the largest and most comprehensive food benchmark to date, with the most significant distinction from current food datasets in the field of spectral food analysis. We believe that spectral analysis is the ultimate approach to achieving multi-dimensional attribute analysis of food. Therefore, we name our benchmark suite as SFOOD (Spectral FOOD).

Through extensive evaluations of cutting-edge methods on SFOOD, we have \textbf{three interesting findings}.
\begin{itemize}
    \item Current LVLMs, even GPT-4o and Gemini, exhibit a relatively poor capability on multi-dimensional food attribute analysis, lagging far behind human.
    \item Compared to the commonly studied categories in ImageNet or COCO, food is perhaps a more challenging topic in computer vision and deserves further investigation.
    \item In contrast to traditional RGB cameras, spectral analysis is crucial for fine-grained food analysis, even with only the same number (3) of spectral bands as RGB.
\end{itemize}

\section{Related Work}
\noindent \textbf{Food datasets.}
Current food datasets can be categorized based on various attributes of food, such as identification, localization, counting, weight and nutrition, recipes, and quality (see Table \ref{dataset_comparison_table}). Due to the strong regional characteristics of food, numerous food recognition datasets have emerged in recent years. AI4Food-NutritionDB \cite{romero2024leveraging} pioneers the work of unifying multiple food datasets to 893 categories and focuses primarily on classification tasks. In terms of localization, given that different food components are often intermingled, there are challenging food segmentation datasets such as UEC Food-256 \cite{kawano2015automatic}, UEC-FoodPix \cite{okamoto2021uec}, and FoodSeg103 \cite{wu2021large}. For counting, public datasets like FSC-147 \cite{ranjan2021learning} and OmniCount-191 \cite{mondal2024omnicount} include a large number of food-related samples. Additionally, there are agricultural plant counting datasets such as RiceSeedlingCount \cite{wu2019automatic}. Regarding weight or nutritional estimation, Nutrition5K \cite{thames2021nutrition5k} proposed by Google serves as an excellent multi-modal dataset. For recipes, Recipe1M+ \cite{marin2021recipe1m+} and its successor RecipeNLG \cite{bien2020recipenlg} have significantly advanced food multi-modal analysis by open-sourcing millions of food images and recipes. In the area of food hyperspectral datasets, HSIFoodIngr-64 \cite{xia2023hsifoodingr} includes 64 ingredients and approximately 3k hyperspectral images, proving spectral analysis significantly boosts food ingredient analysis.
With the emergence of LLM, Food-Sky \cite{zhou2024foodsky} proposes a large-scale food knowledge question and answer dataset in text format. Uni-Food \cite{jiao2024rode} is a large-scale food visual question answering dataset.
Compared to existing datasets, our SFOOD innovatively integrates hyperspectral data in terms of modality and is more comprehensive in food categories and attributes, potentially offering greater value.

\noindent \textbf{RGB-based food analysis.}
Food classification methods \cite{luo2023ingredient,wen2023multi,zhu2023learn} usually design a more appropriate network decomposition for food recognition, and use knowledge distillation and other methods to enhance the generalization and recognition capabilities of the model.
Recent Terrace \cite{nguyen2021terrace} and SibNet \cite{nguyen2022sibnet} employ deep neural networks such as Segment Anything Model (SAM) to address challenges like occlusion and varying shapes and perform simultaneous food counting and segmentation, 
Existing food weight and nutrition analysis algorithms \cite{he2024metafood,tai2023nutritionverse,ruede2021multi} leverage advanced 3D reconstruction techniques and multi-task learning to estimate portion sizes and nutritional content from 2D images, showing promise in improving dietary assessment and nutritional monitoring.
Current food recipe generation algorithms \cite{chhikara2024fire,mohbat2024llava} utilize multi-modal approaches, integrating powerful attention-based vision and language models, to generate comprehensive recipes from food images.
Existing LLM-based approaches \cite{yin2023foodlmm, jiao2024rode} integrate visual and textual data using advanced networks, achieving high accuracy in classification, and employ diverse expert models for complex tasks such as nutrition estimation and recipe generation, enhancing the efficiency and versatility of food-related applications.
Although the development of LVLMs has promoted progress in many vision directions, according to our tests in Table \ref{vqa_result}, current models, even GPT-4o, are still limited in their effectiveness in multi-modal food analysis.

\noindent \textbf{Spectral food analysis.}
Except for the analysis of plants in remote sensing scenarios, there are very few existing methods for hyperspectral food properties analysis. 
Existing hyperspectral food analysis methods \cite{noviyanto2017honey,xia2023hsifoodingr} are characterized by the creation of standardized datasets to utilize the detailed spectral information provided by hyperspectral imaging to enhance tasks such as ingredient prediction and food classification. 
Compared with current methods, we pioneered the use of hyperspectral camera and sugar content analyzers to manually collect and organize challenging food sweetness analysis and Chinese herbal medicine classification tasks to study the need of spectral analysis and band selection \cite{cai2019bs,hu2023one} compared to traditional RGB images.
\section{SFOOD Dataset}\label{dataset_section}

\begin{figure*}[!t]
\centering
\includegraphics[width=1.0\linewidth]{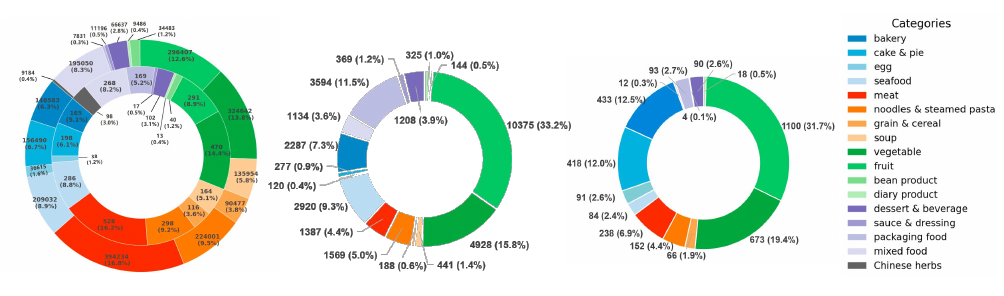}
\caption{\textbf{Distributions of categories} in fine-grained image classification (left), instance segmentation (middle), and object counting (right).
The outer circle of the first figure plots the distribution of the number of images.
}
\label{sfood_distribution}
% \vspace{-4mm}
\end{figure*}

\begin{figure}[!t]
\centering
\includegraphics[width=1.0\linewidth]{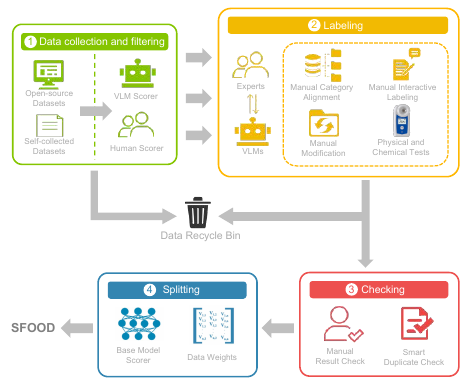}
\caption{\textbf{Data processing procedures of our SFOOD.} }
\label{data_processing_steps}
% \vspace{-4mm}
\end{figure}

In this section, we describe the construction process and data standards of SFOOD. The data processing workflow of SFOOD is depicted in Fig. \ref{data_processing_steps}. To ensure high-quality data, we have divided the dataset processing into four stages: data source and filtering, labeling, checking, and splitting. In the following sections, we first discuss the selection of food categories and then explain each stage of the data processing workflow in detail.

\subsection{Category selection}

We identify the main food categories in SFOOD by following the existing datasets \cite{bossard2014food,minut2021fruits,xu2021multiple} and incorporating the principal food groups established by the United States Department of Agriculture. Finally, we have classified common foods into 17 main categories: bakery, cake \& pie, egg, seafood, meat, noodles \& steamed pasta, grain \& cereal, soup, vegetable, fruit, bean product, dairy product, dessert \& beverage, sauce \& dressing, packaging food, mixed food, and Chinese herbal medicine. Detailed category definitions can be found in the Appendix.
It is worth noting that we have not yet included natural plants and cultivated crops in SFOOD, although they are in some sense a form of food. We plan to add them progressively in future work.

\subsection{Data source}
The data sources are acquired through two approaches: existing open-source datasets and manually collected datasets.

For the existing open-source datasets, we conduct a thorough examination of the categories, distribution, and characteristics of all current food datasets listed in Table \ref{dataset_comparison_table}. Based on a comprehensive assessment of data quality, variability, and relation to 17 main categories, we select 11 datasets as data sources for SFOOD. The 11 datasets are VegFru \cite{hou2017vegfru}, Fruit262 \cite{minut2021fruits}, ChineseFoodNet \cite{hou2017vegfru}, ISIA-500 \cite{min2020isia}, Food-101 \cite{bossard2014food}, VireoFood172 \cite{chen2016deep}, FoodX251 \cite{kaur2019foodx}, Beverage products packaging dataset \cite{chenbeverage}, Biscuit Wrappers Dataset \cite{taspinar2022classification}, Food2K \cite{min2023large}, and CNH-98 \cite{xu2021multiple}. We will continue to collect more datasets to improve SFOOD.

For manually collected data, we mainly collect spectral data as well as food weight and quality data.

\noindent \textbf{Spectral data.}
Using the hyperspectral acquisition equipment, as shown in Fig. \ref{food_capture_device}, we manually collect data for two visually indistinguishable food analysis tasks: food sweetness prediction and herbal medicine classification. For the food sweetness prediction task, we collect data on ten different fruits: sunshine rose grapes, peaches, plums, cherry tomatoes, pears, cherries, waxberries, persimmons, black grapes, and purple grapes. There are approximately 300 sampling points for each fruit. At each sampling point, we use a knife to extract a piece of fruit pulp for juicing and then put it into the sweetness meter for reading and recording the sweetness value. Eventually, we form a dataset of 2,000 samples linking spectral images to sweetness. Additionally, for the herbal medicine classification task, we sample 100 different herbal medicines using the hyperspectral camera. Each type of herbal medicine from various regions will be recorded and photographed several times, forming an average of 600 hyperspectral image samples per class. In total, we collect about 60,000 hyperspectral images and corresponding categories. The size of each hyperspectral image is 35x35x180, where 180 corresponds to the different spectral bands collected by the hyperspectral camera.

\noindent \textbf{The VQA data of food weight and quality.}
We manually collect hundreds of food images and their corresponding weight data based on sophisticated weighing instruments, which results in a VQA dataset for weight estimation. Additionally, for the VQA data related to food quality, we not only source fruit images from publicly available datasets but also take photographs and retrieve from the internet to guarantee a balanced degree of samples in the categories.

\subsection{Data processing}
As shown in Fig. \ref{data_processing_steps}, our data processing consists of four steps: (i) Data source and filtering; (ii) Labeling; (iii) Checking; (iv) Splitting. It is worth noting that our process applies to data processing for various types of food analysis tasks and is expected to inspire the processing of subsequent food datasets.

In the data filtering stage, we employ a scoring system to evaluate target images. For coarse-grained tasks such as estimating the counts or making rough quality judgments, we utilize Large Vision Language Models (LVLMs) and design effective prompts for intelligent judgment and scoring. For tasks that large models are not yet adept at, such as localization and weight estimation, we often resort to manual processing for scoring. Scores range from 1 to 3, symbolizing the difficulty and relevance of the sample; samples with a score of 1, indicating low difficulty or non-compliance with data requirements, are discarded.

In the data annotation stage, we establish different rules for various food analysis tasks, combining physicochemical testing tools for metric measurement. We design different intelligent interactive annotation tools based on open-source AI models to improve annotation efficiency, ultimately relying on experts familiar with the target tasks for manual annotation. For instance, for fine-grained classification tasks, we pre-calculate the features of existing category samples using DINOV2 and automatically compute the similarity of new categories or images to existing ones, providing similar category and image samples for reference to annotators, thereby significantly enhancing annotation efficiency. For instance segmentation tasks, we use the latest interactive segmentation tools to enable annotators to complete the annotation of a complex food object within 2 to 5 clicks.

In the data verification stage, we conduct both manual confirmation of annotation results and the smart duplicate check. The repetitiveness check involves retrieving the top 100 similar samples from the existing dataset using the DINOv2 features of the sample, calculating the SSIM similarity with the current sample, sorting by SSIM similarity, and displaying the Top-10 images on the annotation page. This allows annotators to quickly identify highly repetitive samples, ensuring the high diversity of the dataset.

Finally, in the data partitioning stage, unlike the common random partitioning by proportion, we adopt a weight-adaptive data partitioning method that increases the likelihood of more challenging samples being concentrated in the test set. Take image classification as an example. We first train a neural network $A$ on the complete dataset and then record the confidence $S$ of the correct label predicted by $A$ for each sample. We record $(1.0-S)$ as the difficulty level of the sample. Ultimately, we select samples individually based on the weight of difficulty until 20\% of the data is chosen as test data. Lastly, the remaining 80\% of the data is split in half to form the training and validation sets.

\subsection{SFOOD}
Based on our rigorous data processing workflow, we have organized six tasks to test the performance of current food analysis methods. The organization process is as follows:

\textbf{Fine-grained image classification.} We thoroughly examine existing food-related datasets and, following the guidance of the 17 major categories, filter out subcategories with high data quality that belong to the categories of interest in SFOOD. Through a combination of intelligent and manual annotation, we progressively merge and unify the available categories from 11 datasets. We retain extremely similar yet distinct food categories and merged categories that are almost identical in definition within a single dataset or across different datasets. For example, we differentiate categories in Food2K with repeated English names but different definitions (label IDs are 223, 977). Ultimately, we achieve the largest dataset to date for fine-grained food classification, comprising 3,266 categories.

\textbf{Multi-food instance segmentation.} Due to space limitations, the detailed definition of food instances (and the rules for distinguishing them from components) is placed in the Appendix. We search for high-quality images containing multiple foods from 11 datasets across different major categories. When seeking samples from images of various categories, we use Qwen2-VL-72B as a tool for data filtering to expedite the search for image data. We conduct manual annotations based on interactive instance segmentation tools to significantly accelerate the annotation speed for instance segmentation.

\textbf{Food counting.} Given the high proportion of food in the FSC-147 dataset, we manually select 2k samples. However, these samples are limited in terms of food category diversity. To balance the representation of different major categories of food in the counting dataset, we again use classification datasets and employ Qwen2-VL-72B as a tool to locate approximately 1.5k scene images with a higher number of foods, complex arrangements, and crowdedness within the categories to be supplemented. Subsequently, we manually annotate the newly added images.

\begin{table*}[ht]
\centering
\resizebox{0.8\textwidth}{!}{%
\begin{tabular}{lcccccc|cc|cc}
\hline
\begin{tabular}[c]{@{}l@{}}Model\\ (input 224)\end{tabular} & \begin{tabular}[c]{@{}c@{}}Params\\ (M)\end{tabular} & \multicolumn{1}{l}{GFlops} & \begin{tabular}[c]{@{}c@{}}Top-1\\ (val)\end{tabular} & \begin{tabular}[c]{@{}c@{}}Top-5\\ (val)\end{tabular} & \begin{tabular}[c]{@{}c@{}}Top-1\\ (test)\end{tabular} & \begin{tabular}[c]{@{}c@{}}Top-5\\ (test)\end{tabular} & \begin{tabular}[c]{@{}c@{}}Top-1\\ (ImageNet)\end{tabular} & \begin{tabular}[c]{@{}c@{}}Top-5\\ (ImageNet)\end{tabular} & \multicolumn{1}{c}{\begin{tabular}[c]{@{}c@{}}Top-1\\ (Food-101)\end{tabular}} & \multicolumn{1}{c}{\begin{tabular}[c]{@{}c@{}}Top-5\\ (Food-101)\end{tabular}} \\ \hline
EfficientNet-B0                                             & 8.0                                                 & 0.4                       & 78.24                                                & 93.24                                                & 40.85                                                 & 79.55                                                 & 77.7                                                       & 95.3                                                       & 86.71                                                                          & 95.88                                                                          \\
mixnet-XL                                                   & 11.8                                                & 0.9                        & 79.03                                               & 94.06                                              &     42.23                                             &               82.18                                & 80.5                                                       & 94.9                                                       & 88.62                                                                          & 97.57                                                                          \\
MobileNetV4-L                                               & 36.3                                                & 2.5                        & 80.12                                                & 94.65                                                &    45.34                                             &     84.77                                          & 83.8                                                       & 96.8                                                       & 90.92                                                                          & 98.24                                                                          \\
ResNet-50                                                   & 30.0                                                 & 4.1                        & 79.33                                               & 93.67                                                &   43.56                                             &     83.87                                         & 79.0                                                       & 94.4                                                       & 86.23                                                                          & 97.23                                                                          \\
Swin Base                                                   & 87.7                                                 & 15.2                       &      82.98                                        &     95.86                                       &     \textbf{48.59}                                         &      86.14                                        & 83.5                                                       & 96.5                                                       & 93.91                                                                          & 99.03                                                                          \\
ViT Base                                                    & 86.4                                                 & 16.9                       &      82.37                                            &   95.44                                             &    48.33                                         &           \textbf{86.48 }                                    & 85.5                                                       & N/A                                                        & 93.70                                                                          & 98.98                                                                          \\ \hline
\end{tabular}
}
\caption{\textbf{Comparisons of current methods on food fine-grained classification.} MV4-L denotes MobileNet-V4. The results on ImageNet and Food-101 are provided as a reference to show the difficulty of SFOOD.}
\label{cls_result}
\end{table*}

\begin{table*}[ht]
\centering
\resizebox{0.8\textwidth}{!}{%
\begin{tabular}{llcccccccc|c}
\hline
Model             & Backbone    & \begin{tabular}[c]{@{}c@{}}Params\\ (M)\end{tabular} & \multicolumn{1}{l}{GFlops} & \begin{tabular}[c]{@{}c@{}}mAP\\ (val)\end{tabular} & \begin{tabular}[c]{@{}c@{}}AP50\\ (val)\end{tabular} & \begin{tabular}[c]{@{}c@{}}AP75\\ (val)\end{tabular} & \begin{tabular}[c]{@{}c@{}}mAP\\ (test)\end{tabular} & \begin{tabular}[c]{@{}c@{}}AP50\\ (test)\end{tabular} & \begin{tabular}[c]{@{}c@{}}AP75\\ (test)\end{tabular} & \begin{tabular}[c]{@{}c@{}}mAP\\ (COCO)\end{tabular} \\ \hline
SOLOv2-light      & ResNet-18   & 18.3                                                 & 40.8                       & 26.9                                                    & 34.7                                                     & 29.2                                                     & 14.2                                                     & 19.2                                                     & 15.2         & 29.7                           \\ 

Mask-RCNN         & ResNet-50   & 30.7                                                 & 169.4                      & 28.7                                                & 35.2                                                  & 29.8                                                  & 15.0                                                   & 20.8                                                     & 16.3                       & 35.4                 \\
MS-RCNN           & ResNet-50   & 60.4                                                 & 258.6                      & 12.1                                                    & 16.7                                                     & 13.4                                                     & 6.8                                                     & 9.6                                                      & 7.4              & 36.0                               \\
SOLOv2            & ResNext-101 & 113.9                                                & 319.4                      & \textbf{33.1}                                                    & \textbf{39.4}                                                     & \textbf{34.8}                                                     & \textbf{18.9}                                                    & \textbf{23.5}                                                      & \textbf{20.0}               & 42.4                                  \\

HTC               & ResNet-50   & 76.6                                                 & 1605.4                     & 30.4 & 36.7                                                     & 32.4                                                  & 17.3                                                 & 22.3                                                       &18.6                             & 37.4          \\
Cascade Mask-RCNN & ResNet-101  & 95.5                                                 & 1639.1                     & 29.6                                                   & 35.8                                                    &31.9                                                     & 16.8                                                     & 21.8                                                     & 18.2                 & 37.8                          \\
\hline
\end{tabular}
}
\caption{\textbf{Comparisons of current methods on food instance segmentation.} The mAP results on COCO are provided as a reference to show the difficulty of SFOOD.}
\label{seg_result}
\end{table*}

\textbf{Food visual question answering}. Food VQA is primarily designed to assess the perception capabilities of current LVLMs on foods, divided into six tasks:
\begin{itemize}
    \item Classification. We randomly select 100 data points from the fine-grained classification dataset, using a trained classification model to obtain prediction results, and shuffle the Top-4 categories as question options, with correct answers being those that match the actual category.
    \item Counting. We select 200 data points from the organized food counting dataset, requiring the large model to predict counts based on images, with performance measured by the difference between actual and predicted values.
    \item Weight Estimation. We collect image and weight data for different types of food using weighing instruments (accurate to 0.01g), amassing 200 samples. Performance is assessed by calculating the difference between predicted and actual values.
    \item Quality analysis. We categorize food quality into four aspects, each with several options: maturity (unripe, moderately ripe, fully ripe, overripe), surface condition (no blemish, slight blemish, severe blemish), quality status (fresh, slightly spoiled, severely spoiled), edibility recommendation (directly edible, processed for consumption, inedible). Starting from different labels, images are collected from Fruits-262 and the internet, followed by manual annotation. We judge the final score based on the classification accuracy across the four aspects.
    \item Recipe generation. Although Recipe1M+ contains a vast amount of food-recipe data, many recipes are incomplete or even erroneous. We use LongCLIP to rank the similarity between recipes and their corresponding images on the Recipe1M+ test set, retaining the top 200 most similar recipes as test data. We calculate SacreBLEU and ROUGE-L \cite{chhikara2024fire} to assess the quality of recipe generation.
    \item Nutrition estimation. We randomly extract 100 data points from Nutrition5K for nutrition estimation.

\end{itemize}

\textbf{Spectral fruit sweetness analysis}. For the sweetness prediction task, we sample ten different fruits: sunshine rose grapes, peaches, plums, cherry tomatoes, pears, cherries, waxberries, persimmons, black grapes, and purple grapes, with approximately 300 sampling points per fruit. For each point, we excise a piece of flesh, juice it, and measure the sweetness value using the sweetness meter, resulting in a dataset of 3k samples linking hyperspectral images to actual sweetness values.

\textbf{Spectral herbal classification}. We sample 100 different types of Chinese herbal medicines by recording the category and taking multiple hyperspectral images on different regions, yielding an average of 600 hyperspectral image samples per class, totaling approximately 60k hyperspectral images and corresponding category samples. Each hyperspectral image has a dimension of 35x35x127, where 127 corresponds to the different spectral bands captured by the hyperspectral camera.

The distribution of categories and number of images on the first three tasks are plotted in Fig. \ref{sfood_distribution}.
\section{Experiments}
The experiment consists of six different tasks: fine-grained classification, instance segmentation, counting, multi-modal visual question answering, spectral fruit sweetness analysis, and spectral herbal classification. Since the metrics and experimental details of different tasks are different, we will present them separately. For each set of experiments, we have fixed training settings, but due to space limitations, \textbf{all experimental settings and more detailed discussions are placed in the Appendix}.

\begin{table*}[]
\centering
\resizebox{1.0\textwidth}{!}{%
\begin{tabular}{ccccccccccc}
\hline
                                                                                                                         & \multicolumn{2}{c}{}                                                                                                                     & \textbf{Glm4V-9B} & \textbf{MiniCPM-V-2.6} & \textbf{Qwen2-VL-7B} & \textbf{Qwen2-VL-72B}              & \textbf{Llava1.6-13B} & \textbf{GPT-4o}   & \textbf{Gemini-1.5-flash} & \multicolumn{1}{l}{\textbf{Human}} \\ \hline

\textbf{\begin{tabular}[c]{@{}c@{}}Image classification\\ (100)\end{tabular}}                                             & \multicolumn{2}{c}{Accuracy}                                                                                                             & 0.52              & 0.48                   & 0.46                 & 0.53                               & 0.38                  & {\ul 0.59}        & {\ul 0.59}                & \textbf{0.73}                      \\
\rowcolor[HTML]{DADADA} 
\cellcolor[HTML]{DADADA}                                                                                                 & \multicolumn{2}{c}{\cellcolor[HTML]{DADADA}MAE}                                                                                          & 5.87            & 5.14                 & 4.10               & {\ul 3.18}                        & 7.97                & 3.23            & 7.40                    & \textbf{0.46}                    \\
\rowcolor[HTML]{DADADA} 
\cellcolor[HTML]{DADADA}                                                                                                 & \multicolumn{2}{c}{\cellcolor[HTML]{DADADA}MSE}                                                                                          & 205.7             & 150.2                  & 71.55                & {\ul 23.06}                        & 304.31                & 35.92             & 374.46                    & \textbf{5.08}                      \\
\rowcolor[HTML]{DADADA} 
\multirow{-3}{*}{\cellcolor[HTML]{DADADA}\textbf{\begin{tabular}[c]{@{}c@{}}Counting\\ (200)\end{tabular}}}              & \multicolumn{2}{c}{\cellcolor[HTML]{DADADA}Abnormal cases}                                                                               & 31                & 60                     & 57                   & 82                                 & 47                    & 44                & {\ul 3}                   & \textbf{1}                         \\
                                                                                                                         & \multicolumn{2}{c}{MAE}                                                                                                                  & 0.244             & 0.186                  & 0.284               & 0.139                             & 0.347                & 0.091            & \textbf{0.005}            & {\ul 0.088}                       \\
                                                                                                                         & \multicolumn{2}{c}{MSE}                                                                                                                  & 0.12              & 0.11                   & 0.18                 & 0.04                               & 0.24                  & 0.03              & \textbf{0}                & {\ul 0.02}                         \\
\multirow{-3}{*}{\textbf{\begin{tabular}[c]{@{}c@{}}Weight estimation\\ (200)\end{tabular}}}                             & \multicolumn{2}{c}{Abnormal cases}                                                                                                       & 137               & 176                    & 117                  & {\ul 12}                           & 187                   & 187               & 196                       & \textbf{0}                         \\
\rowcolor[HTML]{DADADA} 
\cellcolor[HTML]{DADADA}                                                                                                 & \cellcolor[HTML]{DADADA}                                                                                                & Accuracy       & 0.50              & 0.62                   & 0.56                 & {\ul 0.67}                         & 0.43                  & 0.50              & 0.64                      & \textbf{0.85}                      \\
\rowcolor[HTML]{DADADA} 
\cellcolor[HTML]{DADADA}                                                                                                 & \cellcolor[HTML]{DADADA}                                                                                                & F1 Score       & 0.49              & 0.62                   & 0.48                 & {\ul 0.68}                         & 0.44                  & 0.50              & 0.59                      & \textbf{0.85}                      \\
\rowcolor[HTML]{DADADA} 
\cellcolor[HTML]{DADADA}                                                                                                 & \multirow{-3}{*}{\cellcolor[HTML]{DADADA}\textbf{Maturity}}                                                             & Recall         & 0.50              & 0.62                   & 0.56                 & {\ul 0.67}                         & 0.43                  & 0.50              & 0.64                      & \textbf{0.85}                      \\
\rowcolor[HTML]{DADADA} 
\cellcolor[HTML]{DADADA}                                                                                                 & \cellcolor[HTML]{DADADA}                                                                                                & Accuracy       & 0.48              & 0.64                   & 0.60                 & {\ul 0.68}                         & 0.40                  & 0.61              & 0.66                      & \textbf{0.92}                      \\
\rowcolor[HTML]{DADADA} 
\cellcolor[HTML]{DADADA}                                                                                                 & \cellcolor[HTML]{DADADA}                                                                                                & F1 Score       & 0.47              & 0.69                   & 0.65                 & {\ul 0.71}                         & 0.39                  & 0.65              & 0.67                      & \textbf{0.92}                      \\
\rowcolor[HTML]{DADADA} 
\cellcolor[HTML]{DADADA}                                                                                                 & \multirow{-3}{*}{\cellcolor[HTML]{DADADA}\textbf{\begin{tabular}[c]{@{}c@{}}Surface \\ Condition\end{tabular}}}         & Recall         & 0.48              & 0.64                   & 0.60                 & {\ul 0.68}                         & 0.40                  & 0.61              & 0.66                      & \textbf{0.92}                      \\
\rowcolor[HTML]{DADADA} 
\cellcolor[HTML]{DADADA}                                                                                                 & \cellcolor[HTML]{DADADA}                                                                                                & Accuracy       & 0.50              & 0.68                   & {\ul 0.85}           & 0.73                               & 0.29                  & 0.67              & 0.77                      & \textbf{0.91}                      \\
\rowcolor[HTML]{DADADA} 
\cellcolor[HTML]{DADADA}                                                                                                 & \cellcolor[HTML]{DADADA}                                                                                                & F1 Score       & 0.59              & 0.73                   & {\ul 0.85}           & 0.77                               & 0.34                  & 0.74              & 0.74                      & \textbf{0.90}                      \\
\rowcolor[HTML]{DADADA} 
\cellcolor[HTML]{DADADA}                                                                                                 & \multirow{-3}{*}{\cellcolor[HTML]{DADADA}\textbf{Health Status}}                                                        & Recall         & 0.50              & 0.68                   & {\ul 0.85}           & 0.73                               & 0.29                  & 0.67              & 0.77                      & \textbf{0.91}                      \\
\rowcolor[HTML]{DADADA} 
\cellcolor[HTML]{DADADA}                                                                                                 & \cellcolor[HTML]{DADADA}                                                                                                & Accuracy       & 0.51              & 0.61                   & 0.54                 & 0.61                               & 0.30                  & 0.54              & {\ul 0.67}                & \textbf{0.87}                      \\
\rowcolor[HTML]{DADADA} 
\cellcolor[HTML]{DADADA}                                                                                                 & \cellcolor[HTML]{DADADA}                                                                                                & F1 Score       & 0.52              & 0.60                   & 0.50                 & 0.62                               & 0.20                  & 0.55              & {\ul 0.64}                & \textbf{0.87}                      \\
\rowcolor[HTML]{DADADA} 
\multirow{-12}{*}{\cellcolor[HTML]{DADADA}\textbf{\begin{tabular}[c]{@{}c@{}}Quality analysis\\ (200)\end{tabular}}}     & \multirow{-3}{*}{\cellcolor[HTML]{DADADA}\textbf{\begin{tabular}[c]{@{}c@{}}Edibility  \\ Recommendation\end{tabular}}} & Recall         & 0.51              & 0.61                   & 0.54                 & 0.61                               & 0.30                  & 0.54              & {\ul 0.67}                & \textbf{0.87}                      \\
                                                                                                                         & \multicolumn{2}{c}{SacreBLEU}                                                                                                            & 3.80              & 2.77                   & 4.85           & 4.37                               & 2.35                  & 3.31              & \textbf{6.21}             &         {\ul 5.72}                   \\
\multirow{-2}{*}{\textbf{\begin{tabular}[c]{@{}c@{}}Recipe generation\\ (200)\end{tabular}}}                             & \multicolumn{2}{c}{ROUGE-L}                                                                                                              & 0.22              & 0.19                   & 0.23           & 0.23 & 0.17                  & 0.21              & \textbf{0.27}             &     {\ul 0.26}                     \\
\rowcolor[HTML]{DADADA} 
\cellcolor[HTML]{DADADA}                                                                                                 & \cellcolor[HTML]{DADADA}                                                                                                & MAE            & 195.49          & 145.56               & 145.88             & {\ul 126.51}                     & 147.33              & \textbf{118.46} & 220.21                  &     154.82                    \\
\rowcolor[HTML]{DADADA} 
\cellcolor[HTML]{DADADA}                                                                                                 & \cellcolor[HTML]{DADADA}                                                                                                & MSE            & 327397.74         & 33030.59               & 39266.37             & {\ul 27660.3}                      & 47521.86              & \textbf{24946.35} & 80758.81                  &     55273.65                     \\
\rowcolor[HTML]{DADADA} 
\cellcolor[HTML]{DADADA}                                                                                                 & \multirow{-3}{*}{\cellcolor[HTML]{DADADA}\textbf{calories}}                                                             & Abnormal cases & 20                & 90                     & {\ul 3}              & \textbf{0}                         & 78                    & 30                & \textbf{0}                &    \textbf{0}                       \\
\rowcolor[HTML]{DADADA} 
\cellcolor[HTML]{DADADA}                                                                                                 & \cellcolor[HTML]{DADADA}                                                                                                & MAE            & 13.05           & 10.16                 & 12.52              & {\ul 8.37}                        & 23.55               & \textbf{6.68}   & 9.68                    &     9.53                   \\
\rowcolor[HTML]{DADADA} 
\cellcolor[HTML]{DADADA}                                                                                                 & \cellcolor[HTML]{DADADA}                                                                                                & MSE            & 204.11            & 157.61                 & 242.22               & {\ul 111.87}                       & 799.13                & \textbf{81.66}    & 153.46                    &    146.91                   \\
\rowcolor[HTML]{DADADA} 
\cellcolor[HTML]{DADADA}                                                                                                 & \multirow{-3}{*}{\cellcolor[HTML]{DADADA}\textbf{fat}}                                                                  & Abnormal cases & 95                & 78                     & 49                   & 2                            & 95                    & 91                & {\ul 1}                &    \textbf{0}                         \\
\rowcolor[HTML]{DADADA} 
\cellcolor[HTML]{DADADA}                                                                                                 & \cellcolor[HTML]{DADADA}                                                                                                & MAE            & 17.60           & 15.28                & 15.08              & {\ul 12.76}                      & 22.45                & \textbf{9.33}   & 24.06                   &    13.66                     \\
\rowcolor[HTML]{DADADA} 
\cellcolor[HTML]{DADADA}                                                                                                 & \cellcolor[HTML]{DADADA}                                                                                                & MSE            & 732.7             & 383.76                 & 437.9                & {\ul 279.95}                       & 1431.26               & \textbf{180.41}   & 925.57                    &    318.34                        \\
\rowcolor[HTML]{DADADA} 
\cellcolor[HTML]{DADADA}                                                                                                 & \multirow{-3}{*}{\cellcolor[HTML]{DADADA}\textbf{carbohydrate}}                                                         & Abnormal cases & 30                & 65                     & 9              & {\ul 2}                         & 64                    & 70                & {\ul 2}                &      \textbf{0}                              \\
\rowcolor[HTML]{DADADA} 
\cellcolor[HTML]{DADADA}                                                                                                 & \cellcolor[HTML]{DADADA}                                                                                                & MAE            & 16.39            & 16.21                & 18.15              & {\ul 11.77}                      & 14.96               & \textbf{7.7279}   & 17.59                     &     13.55                   \\
\rowcolor[HTML]{DADADA} 
\cellcolor[HTML]{DADADA}                                                                                                 & \cellcolor[HTML]{DADADA}                                                                                                & MSE            & 637.59            & 738.24                 & 566.94               & {\ul 301.01}                       & 412.75                & \textbf{147.53}   & 735.22                    &    368.37                        \\
\rowcolor[HTML]{DADADA} 
\multirow{-12}{*}{\cellcolor[HTML]{DADADA}\textbf{\begin{tabular}[c]{@{}c@{}}Nutrition estimation\\ (100)\end{tabular}}} & \multirow{-3}{*}{\cellcolor[HTML]{DADADA}\textbf{protein}}                                                              & Abnormal cases & 29                & 58                     & 9                    & \textbf{0}                         & 69                    & 81                & {\ul 2}                   &      \textbf{0}                   \\ \hline
\end{tabular}%
}
\caption{\textbf{Comparisons of current LVLMs on multi-modal food visual question answering.}}
\label{vqa_result}
\end{table*}

\subsection{Fine-grained food image classification}

\textbf{Metrics.} We compare current methods on both our SFOOD and two popular fine-grained classification datasets: ImageNet-1K and ETH Food-101 \cite{bossard14}. As aforementioned in data processing, SFOOD is divided into 40\%, 40\% and 20\% for training, validation and test set, respectively. Top-1 classification accuracy (Top-1) and Top-5 classification accuracy (Top-5) are adopted as evaluation metrics. 

\textbf{Results.} 
The evaluation results on the 3,266 fine-grained image classifications are shown in Table \ref{cls_result}. Although transformer-based networks such as Swin Transformer and ViT have significantly surpassed human performance on numerous classification tasks, their effectiveness on SFOOD is relatively limited. The Swin Transformer achieves a Top-1 accuracy of 48.59\% on the test set, a significant decrease of 34.9\% compared to the 83.5\% on ImageNet and a notable reduction of 45.3\% compared to the 93.9\% on Food-101, which fully confirms the difficulty of fine-grained food classification. Sample visualizations and more analyses are provided in the Appendix.

\subsection{Food instance segmentation}
\textbf{Metrics.} We compare existing instance segmentation methods, including HTC \cite{chen2019hybrid}, Cascade Mask-RCNN \cite{cai2019cascade}, Mask-RCNN \cite{he2017mask}, MS-RCNN \cite{huang2019mask}, and SOLOv2 \cite{wang2020solov2}. Our primary focus is on instance segmentation metrics, including mean Average Precision (mAP), Average Precision at Intersection over Union (IoU) of 0.5 (AP50), and Average Precision at IoU of 0.75 (AP75).

\textbf{Results.}
Although existing instance segmentation methods have achieved commendable pixel-level localization results on popular instance segmentation datasets such as COCO, their performance on SFOOD is less effective. Based on our evaluations in Table \ref{seg_result}, even computationally intensive methods like HTC and Cascade Mask-RCNN only achieve mAP metrics of 30.4 and 29.6, respectively. Compared to their results on COCO, performances of HTC on mAP decrease from 37.4 to 30.4. Performances of SOLOv2 decrease more significantly from 42.4 to 33.1.
In summary, current instance segmentation methods, regardless of their computational demand and parameter size, exhibit limited performance on the task of food instance segmentation. This suggests that food may be a more challenging subject than current mainstream targets in COCO, urgently requiring more advanced approaches to overcome the challenges of food instance segmentation.

\subsection{Food counting}

\textbf{Metrics.}
We compare existing counting methods, which include LOCA \cite{djukic2023low}, BMNET+ \cite{shi2022represent}, CounTR \cite{liu2022countr}, and RCC \cite{hobley2022learning}. Counting-related metrics containing MAE and Mean Squared Error (MSE) are adopted. For each image, exemplars from three local regions are provided as references.

\textbf{Results.}
In the counting task, as shown in Table \ref{counting_result}, LOCA achieves better counting performances (4.9 MAE on the test set) through iterative prototype adaptation by fusing exemplar shape and appearance information with image features. In contrast, CounTR exhibits weaker learning capabilities for exemplars, resulting in larger errors (7.9 MAE on the test set) in estimated quantities. Furthermore, compared to LOCA, RCC shows a significant gap in MAE between the validation and test sets (3.8 Vs. 6.8), indicating that the generalization capability of RCC is worse than that of LOCA. Overall, the LOCA framework is more effective for food counting.

\subsection{Food visual question answering}

\textbf{Metrics.} The evaluation criteria for food VQA across different dimensions are detailed in Section \ref{dataset_section}. If the target type of answer is not given or the number differs from the standard answer by more than 10 times, the answer is considered abnormal and is not used to compute metrics.

\textbf{Results.} 
As shown in Table \ref{vqa_result}, the food VQA benchmark dataset encompasses six major dimensions, with examples provided in the figure at the first page. We test seven popular LVLMs, both open-source and closed-source, as well as human. We randomly select three non-experts for human evaluation, and the final results are averaged. It is evident that existing large models exhibit poor performance on the first three food-related tasks, with MAE significantly higher than that of the counting model LOCA. Moreover, in weight estimation, the probability of abnormal responses (either not providing a numerical answer or refusing to answer due to uncertainty) is very high, with GPT-4o and Gemini-1.5 reaching 93.5\% and 98\%, respectively. Such a high rate of abnormal answers indicates a weak perception ability of weight in current large vision-language models. Additionally, in terms of food quality perception, human classification accuracy on maturity reaches as high as 85\%, which is significantly higher than the best-performing LVLM, Qwen2-VL-72B, at 67\%. In recipe generation and nutritional estimation, Gemini-1.5 and GPT-4o have shown considerable potential. Overall, Qwen2-VL-72B demonstrates a stronger perception capability for food compared to other LVLMs.

\begin{table}[ht]
\centering
\resizebox{0.85\linewidth}{!}{%
\begin{tabular}{llcccc}
\hline
       & Backbone & \begin{tabular}[c]{@{}c@{}}MAE\\ (val)\end{tabular} & \begin{tabular}[c]{@{}c@{}}MSE\\ (val)\end{tabular} & \begin{tabular}[c]{@{}c@{}}MAE\\ (test)\end{tabular} & \begin{tabular}[c]{@{}c@{}}MSE\\ (test)\end{tabular} \\ \hline
LOCA   & ResNet50 & \textbf{3.004 }                                              & \textbf{34.193 }                                             & \textbf{4.915}                                                & \textbf{105.072}                                              \\
BMNET+ & ConvNets & 4.881                                               & 92.704                                              & 7.632                                                & 224.835                                              \\
CounTR & ViT      & 7.422                                               & 160.356                                             & 7.855                                                & 216.059                                              \\
RCC    & ViT      & 3.771                                               & 69.639                                              & 6.757                                                & 346.891                                              \\ \hline
\end{tabular}
}
\caption{\textbf{Comparisons of current methods on food counting.}}
\label{counting_result}
\end{table}

\begin{table}[ht]
\centering
\resizebox{1.0\linewidth}{!}{%
\begin{tabular}{lcccccccc}
\hline
\multirow{2}{*}{Model} & \multicolumn{2}{c}{RGB} & \multicolumn{2}{c}{PCA (40)} & \multicolumn{2}{c}{PCA (20)} & \multicolumn{2}{c}{\begin{tabular}[c]{@{}c@{}}Selected (3)\\ (944, 1013, 1019 nm)\end{tabular}} \\
                       & MAE        & MSE        & MAE           & MSE          & MAE           & MSE          & MAE                                       & MSE                                       \\ \hline
SSFTT                  & 1.722      & 4.940       & 1.336         & 3.214        & 1.481         & 3.742        & 1.482                                     & 3.695                                     \\
ResNet-50              & 1.608      & 4.606      & 1.270          & 3.069        & 1.287         & 3.026        &  1.371                                &   3.310                               \\
MV4-L          & 1.931      & 7.837      & 1.522         & 4.744        & 1.562         & 4.830         &          1.629                      &       5.287                           \\ \hline
\end{tabular}
}
\caption{\textbf{Comparisons of different combinations of wavelengths on fruit sweetness prediction.}}
\label{sweetness_result}
\end{table}

\subsection{Spectral fruit sweetness analysis}

\textbf{Metrics.} Brix is a measure of the sugar content in a food item. For the task of sweetness analysis, we assess the performance of sweetness prediction based on the error in Brix values. We obtain RGB images by synthesizing images at three wavelengths: 698.07 nm, 544.14 nm, and 436.53 nm. Similarly, for wavelengths selected by Principal Component Analysis (PCA) or band selection, we synthesize the corresponding multispectral images.

\textbf{Results.}
As shown in Table \ref{sweetness_result}, when training neural networks to learn and analyze sweetness based on RGB wavelengths, ResNet-50 achieves a lower MAE than other models. When analyzing based on 40-channel multispectral images generated after PCA, the MAE significantly decreases from 1.608 to 1.270, demonstrating the necessity of spectral data for sweetness analysis. After employing BSNet for band selection, empowered by three selected bands, SSFTT greatly surpassed its results on RGB images (1.482 vs 1.722 on Brix), even matching the performances with 20 channels (1.481). Similar phenomena were observed in ResNet-50 and MobileNetV4-L. Therefore, we arrive at two conclusions: (1) Spectral data is crucial for food property analysis; (2) Analysis based on three prominent spectral bands can significantly outperform RGB-based analysis.

\begin{table}[ht]
\centering
\resizebox{1.0\linewidth}{!}{%
\begin{tabular}{lllllllcc}
\hline
\multirow{2}{*}{Model} & \multicolumn{2}{c}{RGB (3)}                           & \multicolumn{2}{c}{PCA (40)}                          & \multicolumn{2}{c}{PCA (20)}                          & \multicolumn{2}{c}{\begin{tabular}[c]{@{}c@{}}Selected (3)\\ (822, 903, 1007 nm)\end{tabular}} \\
                       & \multicolumn{1}{c}{Top-1} & \multicolumn{1}{c}{Top-5} & \multicolumn{1}{c}{Top-1} & \multicolumn{1}{c}{Top-5} & \multicolumn{1}{c}{Top-1} & \multicolumn{1}{c}{Top-5} & Top-1                                     & Top-5                                     \\ \hline
SSFTT                  & 32.90                     & 63.89                     & 70.98                     & 90.60                     & 70.46                     & 91.83                     & 45.30                                        & 77.80                                    \\
ResNet-50              & 29.28                     & 60.87                     & 64.16                     & 90.94                     & 63.48                     & 89.97                     & 32.60                                       & 70.10                                     \\
MV4-L          & 14.77                     & 40.32                     & 17.23                     & 38.38                     & 15.59                     & 36.22                     & 15.90                                   & 33.20                                    \\ \hline
\end{tabular}
}
\caption{\textbf{Comparisons of different combinations of wavelengths on spectral herbal classification.}}
\label{herbal_result}
\end{table}

\subsection{Spectral herbal classification}

\textbf{Metrics.} Top-1 classification accuracy (Top-1) and Top-5 classification accuracy (Top-5) are adopted as metrics. 

\textbf{Results.} 
Due to the visually similar appearances of Chinese herbal medicine (see the figure in the first page), classification based on RGB images is extremely challenging (29.28\% Top-1 accuracy for ResNet-50). As shown in Table \ref{herbal_result}, this difficulty is greatly alleviated with the introduction of additional spectral data, with the Top-1 accuracy of SSFTT improving dramatically from 32.9\% to 70.98\% (based on 40 bands). Notably, when prominent spectral bands that represent the characteristic spectrum of herbals are selected, SSFTT surpasses the analytical precision based on RGB images by large margins (45.30 VS. 32.90). This huge improvement indicates that the application prospects of research combining spectral, food, and AI are substantial, capable of addressing numerous practical challenges.

\section{Conclusion}
In this paper, we introduce the SFOOD benchmark suite, which includes over 3,266 food categories and 2,351 k samples, making it the largest and most comprehensive food dataset to date, significantly advancing the development of multi-dimensional food analysis. Our evaluations demonstrate that food analysis remains challenging for current models, highlighting the importance of spectral data for accurate food attribute analysis. The comprehensive nature of SFOOD and its spectral analysis benchmarks position it as a valuable resource for future research, and we will continue to expand it to support advancements in multi-dimensional food analysis.
{
    \small
    \bibliographystyle{ieeenat_fullname}
    \bibliography{main}
}

\clearpage
\setcounter{page}{1}
\maketitlesupplementary

\begin{figure*}[!t]
\centering
\includegraphics[width=1.0\linewidth]{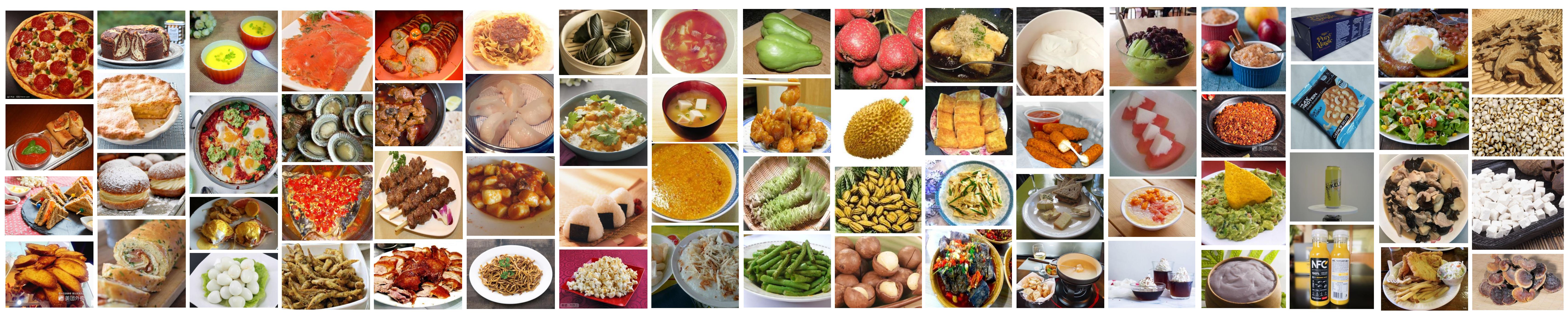}
\caption{Four example subcategories in each major category. From left to right: bakery, cake \& pie, egg, seafood, meat, noodles \& steamed pasta, grain \& cereal, soup, vegetable, fruit, bean product, dairy product, dessert \& beverage, sauce \& dressing, packaging food, mixed food, and Chinese herbal medicine.
}
\label{category-sample}
% \vspace{-4mm}
\end{figure*}

\section{The definition of food categories}

We have classified common foodstuffs, including raw ingredients and processed dishes, into 17 major categories: bakery, cake \& pie, egg, seafood, meat, noodles \& steamed pasta, grain \& cereal, soup, vegetable, fruit, bean product, dairy product, dessert \& beverage, sauce \& dressing, packaging food, mixed food, and Chinese herbal medicine. It is worth noting that these categories do not represent all foods. For example, many plants are edible, but they are not included in the SFOOD dataset for the time being. We will continue to expand the categories of SFOOD in the future.

We explain the definitions of food categories in our dataset as follows and list a few representative samples of each category in Fig. \ref{category-sample}.

\begin{itemize}
    \item \textbf{Bakery}. Foods primarily made from flour, which are produced through fermentation and baking processes, typically requiring high temperatures and specific baking times to complete. They can serve as staple foods or snacks.
    \item \textbf{Cake \& Pie}. Foods  made from flour and often sugar, eggs, butter,etc., formed through baking. They are less commonly used as a staple food.
    \item \textbf{Dessert \& Beverage}. Sweet foods and drinks, without flour, made in various ways such as freezing, stirring, mixing.It can range from liquid to solid textures.
    \item \textbf{Egg}. Eggs including chicken, duck, and quail eggs, prepared in various ways.
    \item \textbf{Seafood}. Animals the are living underwater, including freshwater fish, marine fish, shrimp, crabs, and scallops.
    \item \textbf{Meat}. Dishes that are primarily meat-based, including pork, beef, lamb, and poultry such as chicken and duck.
    \item \textbf{Noodles \& Steamed Pasta}. Flour-based foods that do not require baking.
    \item \textbf{Grain \& Cereal}. Rice and other grain-based dishes, served as staple food in Asia frequently.
    \item \textbf{Soup}. Dishes that contain a large amount of liquid, including soup and porridge, made with diverse ingredients.
    \item \textbf{Vegetable}. Vegetables include raw and processed vegetable dishes.
    \item \textbf{Fruit}. Fruits and nuts, mostly consumed directly, and some processed like salads or fruit mixes.
    \item \textbf{Dairy product}. Foods primarily made from milk.
    \item \textbf{Bean product}. Foods primarily made from bean.
    \item \textbf{Sauce \& Dressing}. Dipping sauces and jams.
    \item \textbf{Packaging food}. Foods with package, including those in boxes, bottles, bags, and cans.
    \item \textbf{Mixed food}. Dishes with a balanced mix of ingredients that cannot be categorized under meat, rice, vegetables or any single category.
    \item \textbf{Chinese herbal medicine}. Processed medicinal materials from herbs.
\end{itemize}

Because of the huge amount of our categories, the detailed food subcategories with their parent category and total image number are listed in our github repository.

\section{The definition of food instances}

Here, we discuss the annotation standards for food instance segmentation separately.

After examining the labeling rules of current food datasets such as UEC Food-256, UEC-FoodPix, and FoodSeg103, we find that the instance labeling rules vary significantly between datasets. Given the often unclear distinctions between food and ingredients, and the typically crowded environment of food items, the unification of food instance labeling rules becomes crucial for research in multi-food localization.

Through continuous corrections and discussions during long-term manual annotation process, we have developed an effective method for distinguishing between foods and ingredients, distilling a practical standard for food instance definition that is applicable to foods from different regions. A food instance typically possesses the following characteristics:

\begin{itemize}
    \item Atomicity of food. If a component, when separated, renders the entire food item definitionally incomplete, it is considered the smallest indivisible unit of that food. If a component is annotated as a separate food item (e.g., decorative garnish), that area should not be considered a standalone food. It should be regarded as an accessory to the food it accompanies and annotated together with its associated food, not as an independent item.
    \item Occlusion of food. Regions and pixels obscured by other foods or foreign objects should not be annotated as food foreground. However, if obscured by food-related tableware, it should be annotated because the tableware possesses food-related attributes, and the food should not be considered obscured due to the presence of tableware.
    \item Independence of food. In three-dimensional space, food items with one dimension smaller than 2 centimeters (such as rice or beans) should be annotated as a group when clustered. Because of their small size, they are unlikely to be consumed individually, and thus single items often lack independence. Non-rigid foods (such as fries) also lack independence and should be annotated as a group rather than individually when intermingled with other similar items.
\end{itemize}

Our dataset has 6066 images for multi-food instance segmentation task in total, including 2427 for training, 2426 for validation and 1213 for testing.

\begin{table*}[!]
\centering
\begin{tabular}{|cc|cc|cc|cc|}
\hline
id & wavelength(nm) & id & wavelength(nm) & id & wavelength(nm) & id  & wavelength(nm) \\
\hline
1  & 370.74         & 33 & 555.23         & 65 & 730.77         & 97  & 897.87         \\
2  & 376.88         & 34 & 560.77         & 66 & 736.2          & 98  & 902.7          \\
3  & 382.99         & 35 & 566.3          & 67 & 741.62         & 99  & 907.49         \\
4  & 389.06         & 36 & 571.82         & 68 & 747.04         & 100 & 912.23         \\
5  & 395.1          & 37 & 577.34         & 69 & 752.46         & 101 & 916.94         \\
6  & 401.1          & 38 & 582.86         & 70 & 757.86         & 102 & 921.61         \\
7  & 407.08         & 39 & 588.37         & 71 & 763.26         & 103 & 926.23         \\
8  & 413.02         & 40 & 593.88         & 72 & 768.64         & 104 & 930.8          \\
9  & 418.94         & 41 & 599.39         & 73 & 774.02         & 105 & 935.33         \\
10 & 424.83         & 42 & 604.89         & 74 & 779.39         & 106 & 939.81         \\
11 & 430.69         & 43 & 610.39         & 75 & 784.74         & 107 & 944.24         \\
12 & 436.53         & 44 & 615.88         & 76 & 790.08         & 108 & 948.62         \\
13 & 442.34         & 45 & 621.38         & 77 & 795.41         & 109 & 952.94         \\
14 & 448.14         & 46 & 626.87         & 78 & 800.73         & 110 & 957.2          \\
15 & 453.91         & 47 & 632.36         & 79 & 806.04         & 111 & 961.41         \\
16 & 459.66         & 48 & 637.85         & 80 & 811.32         & 112 & 965.56         \\
17 & 465.39         & 49 & 643.33         & 81 & 816.59         & 113 & 969.64         \\
18 & 471.11         & 50 & 648.82         & 82 & 821.85         & 114 & 973.66         \\
19 & 476.8          & 51 & 654.3          & 83 & 827.09         & 115 & 977.62         \\
20 & 482.48         & 52 & 659.78         & 84 & 832.3          & 116 & 981.51         \\
21 & 488.15         & 53 & 665.26         & 85 & 837.5          & 117 & 985.33         \\
22 & 493.8          & 54 & 670.73         & 86 & 842.68         & 118 & 989.08         \\
23 & 499.44         & 55 & 676.21         & 87 & 847.83         & 119 & 992.75         \\
24 & 505.06         & 56 & 681.68         & 88 & 852.96         & 120 & 996.35         \\
25 & 510.67         & 57 & 687.15         & 89 & 858.06         & 121 & 999.87         \\
26 & 516.28         & 58 & 692.61         & 90 & 863.14         & 122 & 1003.31        \\
27 & 521.87         & 59 & 698.07         & 91 & 868.2          & 123 & 1006.66        \\
28 & 527.45         & 60 & 703.53         & 92 & 873.22         & 124 & 1009.93        \\
29 & 533.02         & 61 & 708.99         & 93 & 878.21         & 125 & 1013.12        \\
30 & 538.58         & 62 & 714.44         & 94 & 883.18         & 126 & 1016.22        \\
31 & 544.14         & 63 & 719.89         & 95 & 888.11         & 127 & 1019.22        \\
32 & 549.69         & 64 & 725.33         & 96 & 893.01         &     &    \\
\hline

\end{tabular}

\caption{Wavelength of hyperspectral camera}
\label{hyperspectral camera wavelength}

\end{table*}

\section{Hyperspectral Data}

With the use of dual hyperspectral cameras(Fig. \ref{food_capture_device}), we have obtained spectral data with wavelength between 400 nm to 1020 nm shown in Table \ref{hyperspectral camera wavelength}.

\section{Experiments Setting and Analysis}

All experiments are conducted on a server with 8 NVIDIA GeForce RTX 3090.

\subsection{Fine-grained image classification}

To fairly evaluate each model, we employ pre-trained models from ImageNet and their training settings suggested by the Timm repository. The input resolution was set to 224x224 pixels, and the training takes 150 epochs for all models. A detailed overview of the training configurations for each model is presented in Table \ref{setting_classification}.

\begin{table*}[]
\resizebox{1.0\linewidth}{!}{%
\begin{tabular}{|c|ccccccc|c|}
\hline
model           & lr       & lr decay & weight decay rate & optimizer & dropout & momentum & other                                                  \\
\hline
ResNet-50       & 0.05     & cosine   & 2.00e-05          & sgd       & 0       & 0.9      &    None                                              \\
EfficientNet-B0 & 1.00e-05 & step     & 1.00e-05          & rmsproptf & 0.2     & 0.9      &    None                                                \\
MixNet-XL       & 0.256    & step     & 1.00e-05          & rmsprop   & 0       & 0.9      &    None                                                \\
MobileNetV4-L   & 0.01     & cosine   & 2.00e-05          & sgd       & 0.2     & 0.9      & mixup=0.8,cutmix=1.0,mixupprob=0.3,mixupswitchprob=0.5 \\
ViT Base        & 2.00e-05 & step     & 0.03              & sgd       & 0.1     & 0.9      & patch=16                                               \\
Swin Base       & 2.00e-04 & cosine   & 1.00e-08          & sgd       & 0       & 0.9      & patch=4, windows=7,dropout\_path=0.2   \\
\hline
\end{tabular}
}
\caption{Training settings for fine-grained classification task.}
\label{setting_classification}
\end{table*}

\begin{figure*}[!t]
\centering
\includegraphics[width=0.8\linewidth]{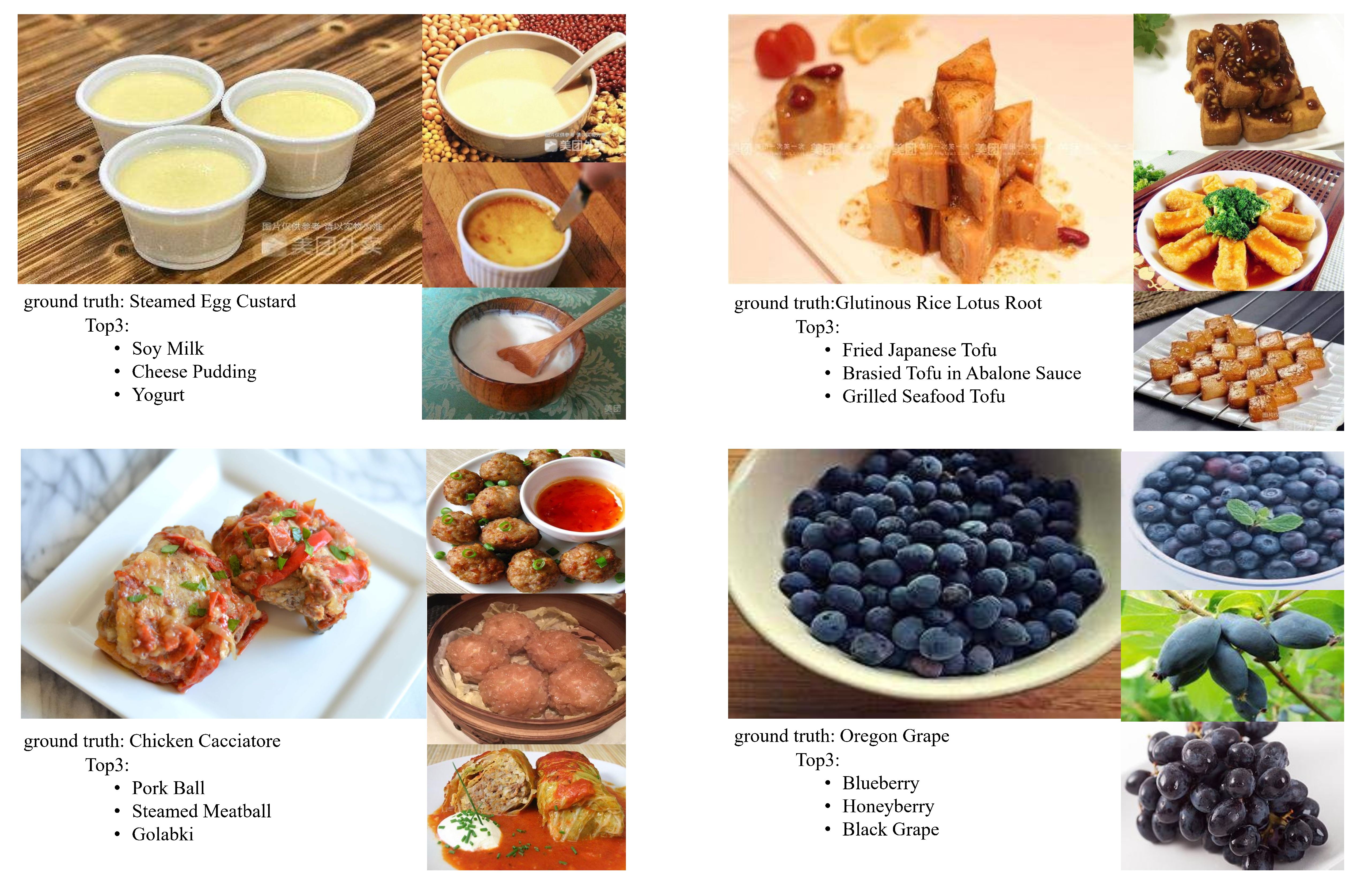}
\caption{Fine-grained classification results by Swin Transformer on the test set.}
\label{classification-vis}
% \vspace{-4mm}
\end{figure*}

The experimental outcomes on the test set indicate that while the majority of cases can be correctly classified into parent categories, the Top1 accuracy rate still remains unsatisfactory. 
SFOOD, a integration of various datasets, has introduced dishes with similar appearances across different datasets to make itself comprehensive and international. 
Examples of such confounding dishes include egg custard, cheese pudding, and soy milk are shown in Fig. \ref{classification-vis}, which raise the demand for precise fine-grained classification.

\subsection{Multi-food instance segmentation}

For a balanced evaluation, we utilize pre-trained models on the COCO dataset and the recommended training/fine-tuning configurations release by the MMdetection repository to ensure reasonable performances. The training process span 40 epochs, with the specific settings for each model delineated in Table \ref{setting_segmentation}.

\begin{table}[]
\resizebox{1.0\linewidth}{!}{%
\begin{tabular}{|c|ccccc|}
\hline
model             & backbone    & lr   & momentum & optimizer & weight decay \\
\hline
SOLOv2-light      & ResNet-18   & 0.01 & 0.9      & sgd       & 1.00e-04     \\
Mask-RCNN         & ResNet-50   & 0.08 & 0.9      & sgd       & 4.00e-05     \\
MS-RCNN           & ResNet-50   & 0.02 & 0.9      & sgd       & 1.00e-04     \\
SOLOv2            & ResNext-101 & 0.01 & 0.9      & sgd       & 1.00e-04     \\
HTC               & ResNet-50   & 0.02 & 0.9      & sgd       & 1.00e-04     \\
Cascade Mask-RCNN & ResNet-101  & 0.02 & 0.9      & sgd       & 1.00e-04    \\
\hline
\end{tabular}
}
\caption{Training settings for multi-food instance segmentation task.}
\label{setting_segmentation}
\end{table}

\begin{figure*}[!t]
\centering
\includegraphics[width=0.9\linewidth]{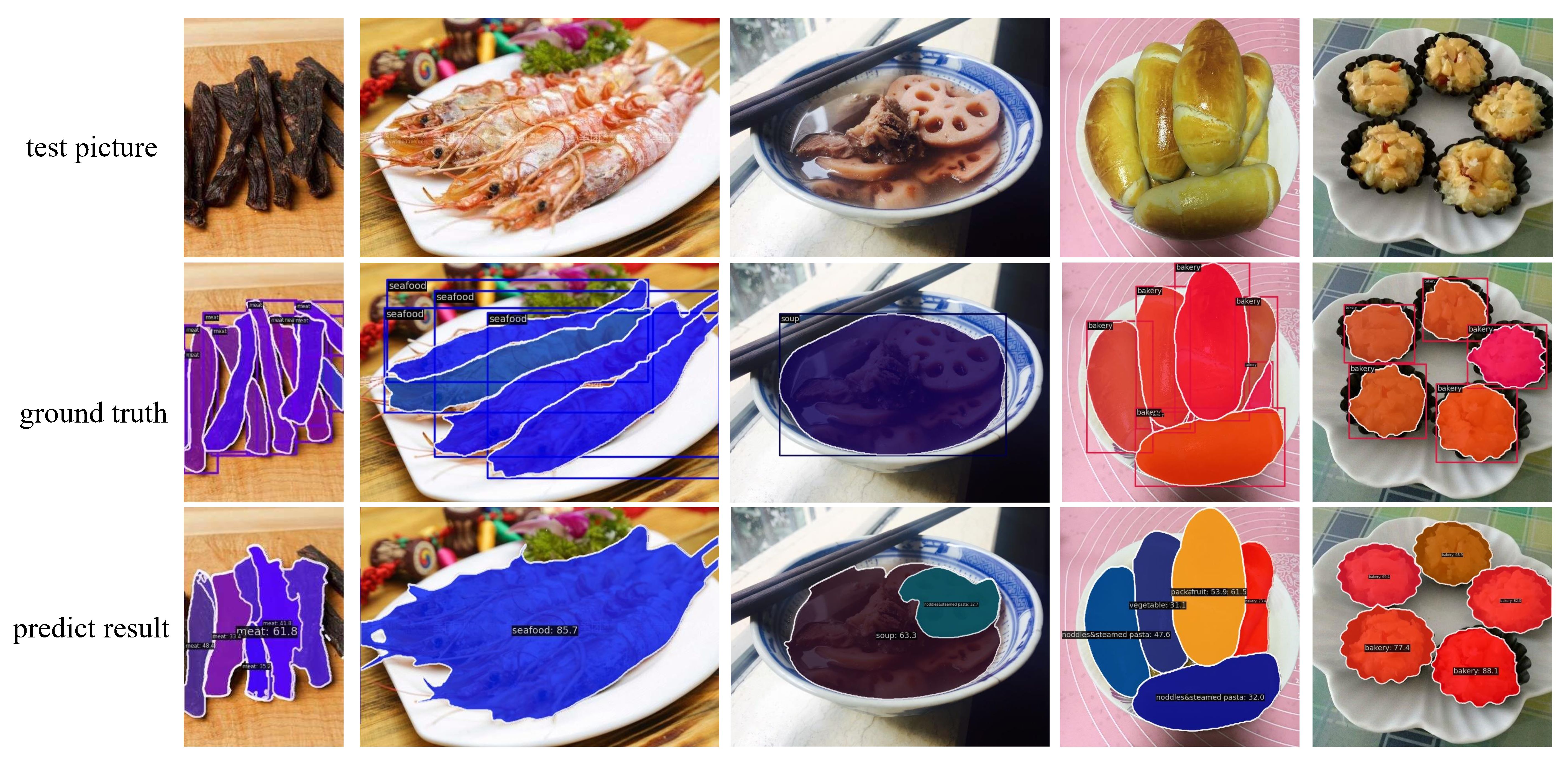}
\caption{Multi-food instance segmentation results by SOLOv2 on the test set.}
\label{segmentation-vis}
% \vspace{-4mm}
\end{figure*}

Through visual analysis of bad cases in Fig. \ref{segmentation-vis}, it is evident that the concept of instances is still not distinguished well in some scenarios, such as in the case of grilled skewers, ingredients in soup and food with challenging placement. Secondly, due to the diversity of food items, it is hard to classify major categories of food clearly in instance segmentation task, primarily because there are foods with similar appearances across different categories, such as bread and pasta. Additionally, there remains difficulty in the separation of food from its containers.

\subsection{Food counting}

Employing a hybrid dataset comprising FSC-147 and self-labeled data, we train our object counting models in accordance with the training guidelines following previous works. The training duration is set to 200 epochs, with the detailed training parameters for each model outlined in Table \ref{setting_counting}.

\begin{table}[]
\resizebox{1.0\linewidth}{!}{%
\begin{tabular}{|c|ccccc|c|}
\hline
model  & backbone   & lr       & weight decay & dropout & optimizer & other                            \\
\hline
LOCA   & RestNet-50 & 1.00e-04 & 1.00e-04     & 0.1     & AdamW     &                                  \\
BMNET+ & ConvNets   & 1.00e-05 & 5.00e-04     & 0       & AdamW     &                                  \\
CounTR & ViT        & 2.00e-04 & 0.05         & 0       & AdamW     & pre-trained on FSC for 300 epochs \\
RCC    & ViT        & 3.00e-05 & 0            & 0       & Adam     &                          \\
\hline
\end{tabular}
}
\caption{Training settings for food counting task.}
\label{setting_counting}
\end{table}

\begin{figure}[!t]
\centering
\includegraphics[width=0.8\linewidth]{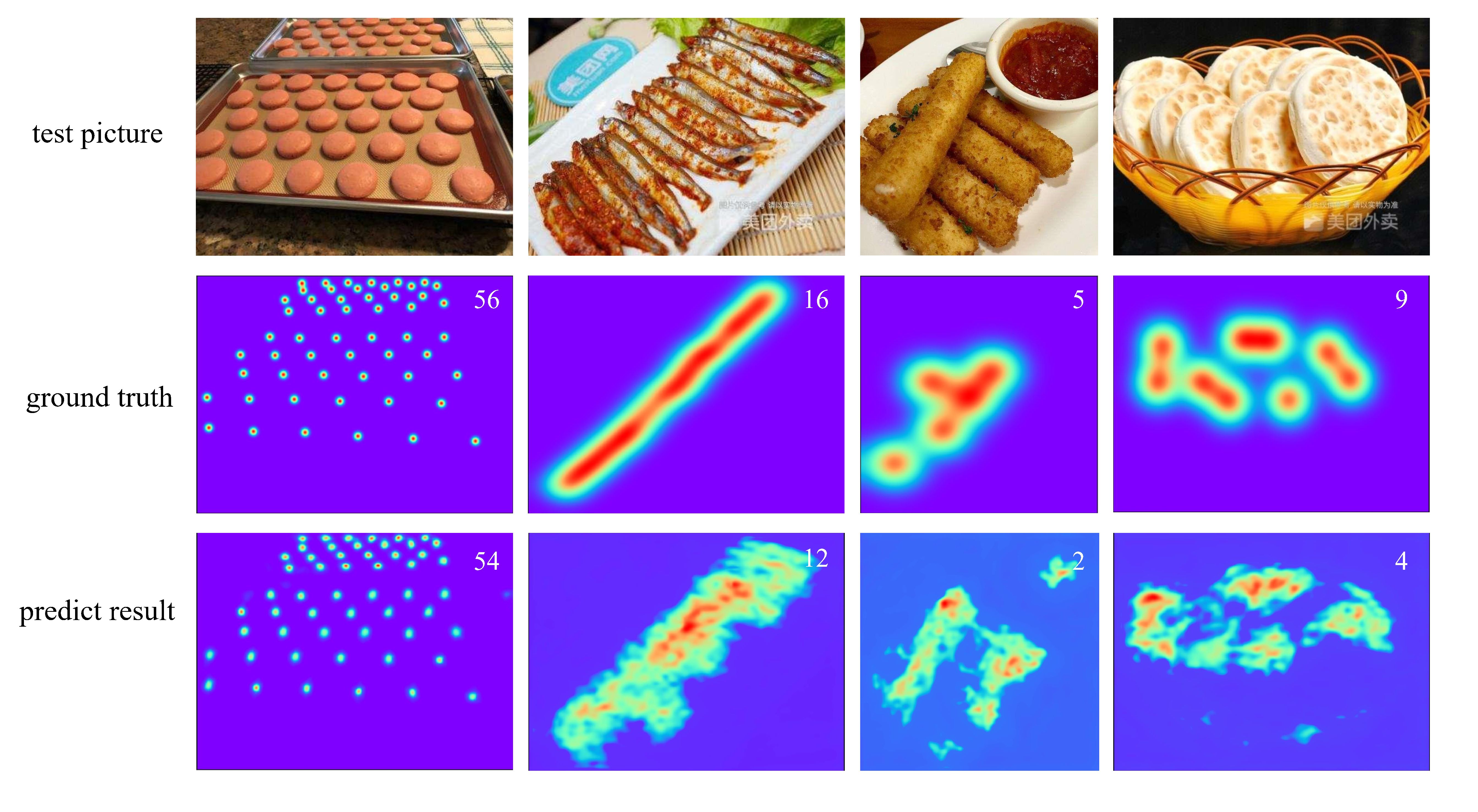}
\caption{Food counting results by LOCA.}
\label{counting-vis}
% \vspace{-4mm}
\end{figure}

In real-world food-related condition, tasks involving the counting of hundreds or thousands of objects are rare. Through evaluations, we find that, while narrowing the value range might seem to reduce task difficulty, challenges persist even in scenarios with a small number of objects, such as visual prompt with various appearance, and instances where objects are covered by jam or sauce (see Fig. \ref{counting-vis}).

\subsection{Food visual question answering}

We select several samples in each VQA task and three models with relatively capable performance, MiniCMP-V-2.6, Qwen2-VL-72B,GPT-4o. We select some sample from testing and highlight the reasonable details in green, the abnormal part in red, and the refusal in blue.

\begin{itemize}
    \item Image classification: For rare species (e.g., Turnip Cabbage) and fine distinctions between similar categories (e.g., Arctic bay sashimi and tuna sashimi), Qwen2-VL-72B outperforms GPT-4o.
    \item Food counting: All models exhibit errors and rarely count the number accurately while GPT-4o counts more close to the correct answer in general. They tend to refuse to count the exact number in situations with a large quantity and make mistakes even with small quantities when the arrangement is challenging.
    \item Weight estimation: Most images, including dishes and raw fruits and vegetables, receive a refusal from MiniCMP-V-2.6 and GPT-4o, except for common fruits like bananas and apples, while Qwen2-VL-72B answered directly. It is worth noting that the weight of some packaging foods is obtained through OCR capabilities, as the prediction is the same as the weight on the package.
    \item Quality analysis: The performance of the three models is similar, with higher accuracy in surface condition and health status analysis, and lower accuracy in more abstract and comprehensive concepts such as ripeness and recommendation for consumption.
    \item Recipe generation: The models add some details, providing ranges for time and temperature. The main differences between model responses and the reference answers occur in the fine-grained recognition of ingredients, such as foods with similar appearances (meat and eggs) and different cooking methods for the same food (steaming and roasting), highlighting the significance of fine-grained food classification.
    \item Nutrition estimation: GPT-4o typically list the nutrient content corresponding to the recognized foods, meeting the precision requirement of six decimal places mostly and providing more convincing predictions compared to the other two models. Qwen2-VL-72B directly response to all questions. However, some of its responses do not meet the output format requirement. MiniCPM-V-2.6 refuses to answer most questions, with many anomalies present in the predicted values.
\end{itemize}

\begin{figure}[!t]
\centering
\includegraphics[width=1\linewidth]{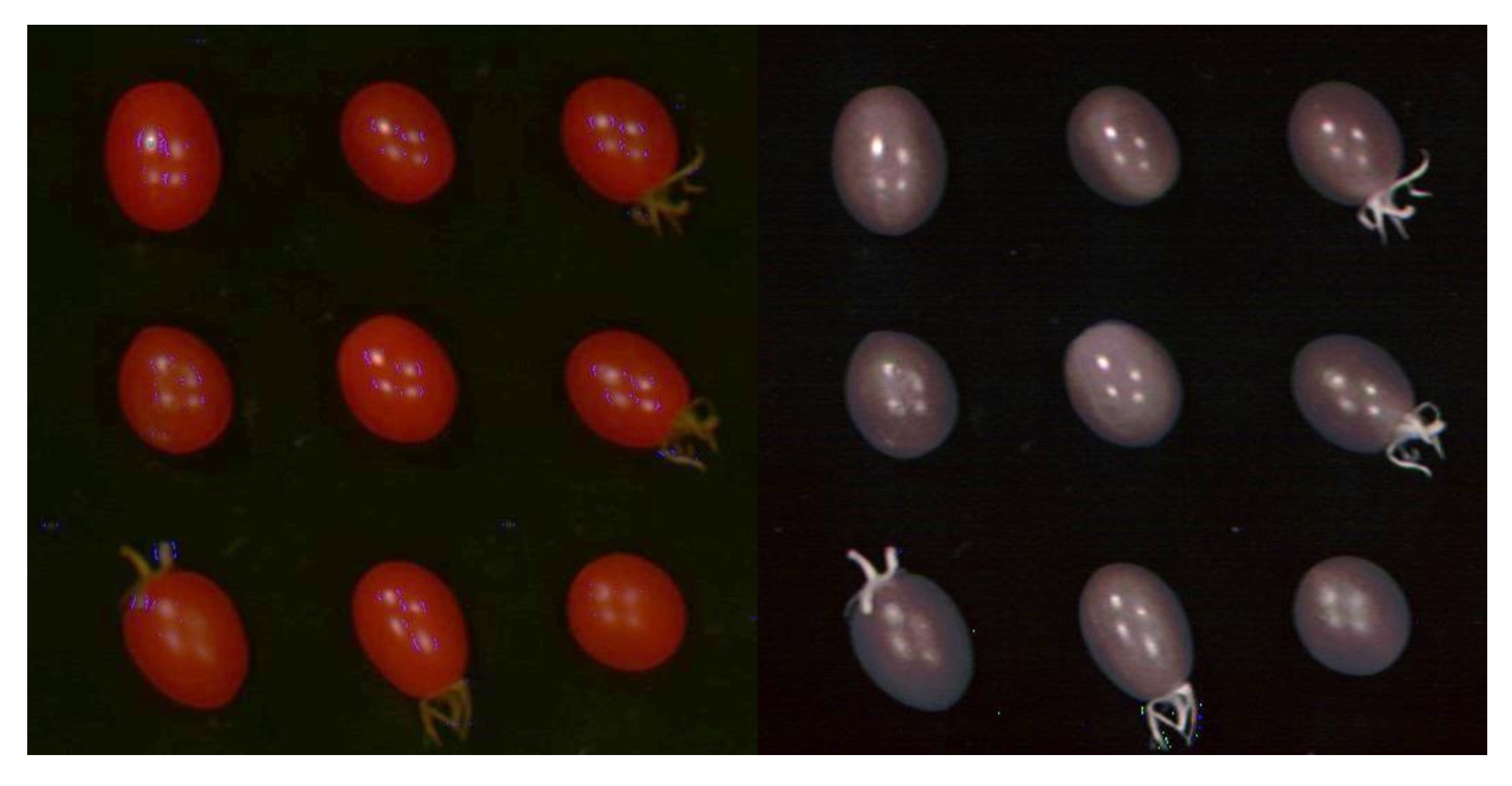}
\caption{Tomatoes with different sweetness shown in RGB (left) and 3-wavelengths spectral images selected by BSNet (right), while the one in the center with the highest brix value.}
\label{spectral_brix}
% \vspace{-4mm}
\end{figure}

\subsection{Spectral fruit sweetness analysis}

For the prediction of sweetness, we present fruits with varying Brix levels in RGB images and 3-wavelengths spectral images selected by BSNet (Fig. \ref{spectral_brix}). As an intrinsic attribute of fruit, we human struggle to discern sweetness levels solely from the appearance in RGB images. However, the NIR bands provide valuable insights beyond the visible spectrum. When analyzing based on 40-channel multispectral images generated after PCA, the MAE significantly decreases from 1.608 to 1.270, demonstrating the necessity of spectral data for sweetness analysis.
By employing a powerful band selection technique, we can select a limited number (3 to 5) of bands, far less than the 127, to achieve highly accurate Brix predictions.

\subsection{Spectral herbal classification}

\begin{figure}[!t]
\centering
\includegraphics[width=1\linewidth]{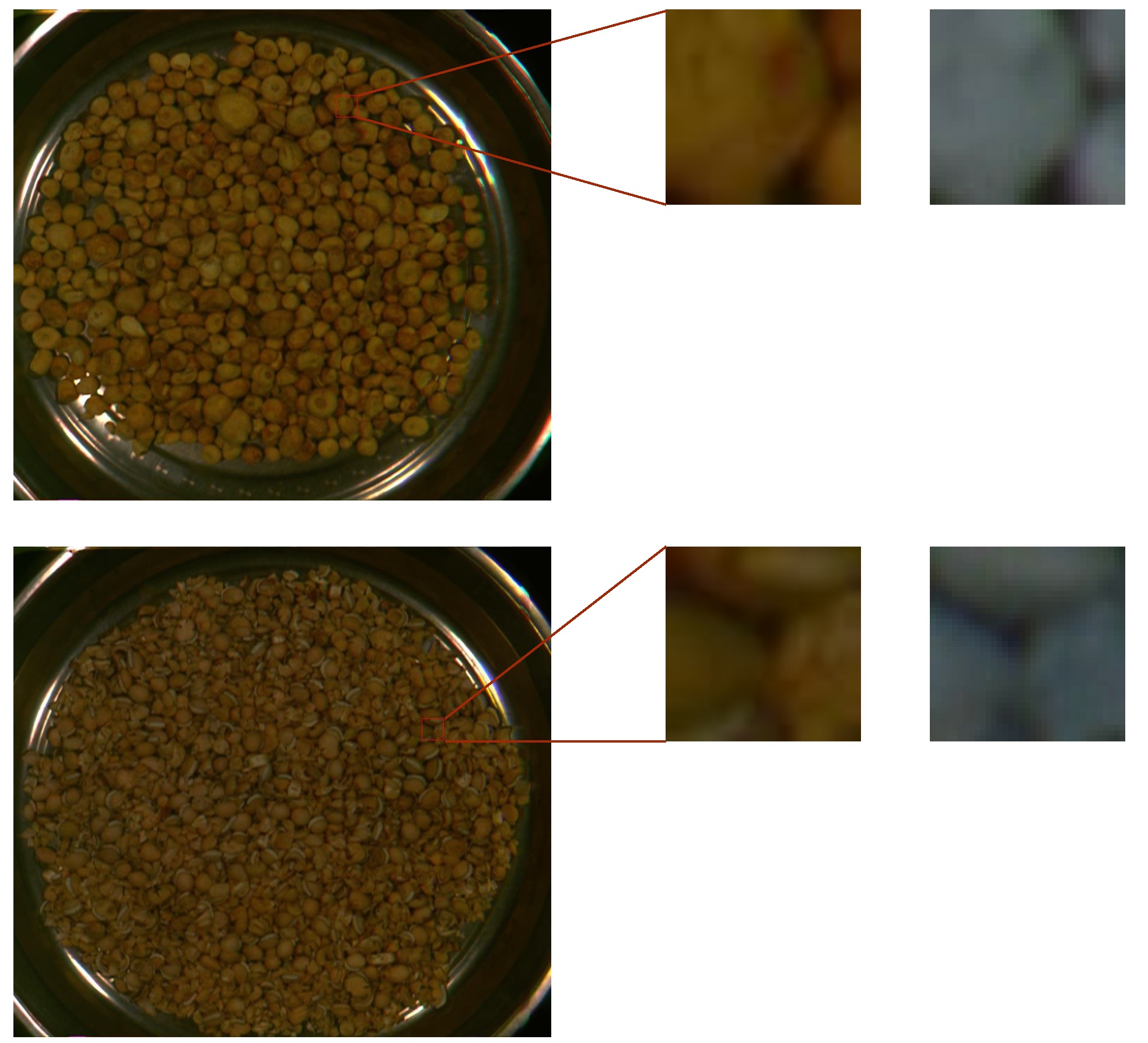}
\caption{A 35x35 crop from \textbf{Processed Ternate Pinellia} and \textbf{White Hyacinth Bean} shown in RGB (middle) and 3-wavelengths spectral images selected by BSNet(right). 
}
\label{spectral_herbs}
% \vspace{-4mm}
\end{figure}

It is quite challenging to recognize the Chinese herbal medicine from only 35x35 pixels, which means the shape features can hardly be learned for recognition.  As depicted in Fig. \ref{spectral_herbs}, we illustrate two similar categories using RGB and 3-wavelengths selected by BSNet to highlight the similarities between these categories. It is worth noting that, due to the visually similar appearances of Chinese herbal medicine (see the figure in the first page), classification based on RGB images is extremely challenging (29.28\% Top-1 accuracy for ResNet-50). As shown in Table \ref{herbal_result}, this difficulty is greatly alleviated with the introduction of additional spectral data, with the Top-1 accuracy of SSFTT improving dramatically from 32.9\% to 70.98\% (based on 40 bands).

\begin{figure*}[!t]
\centering
\includegraphics[width=0.8\linewidth]{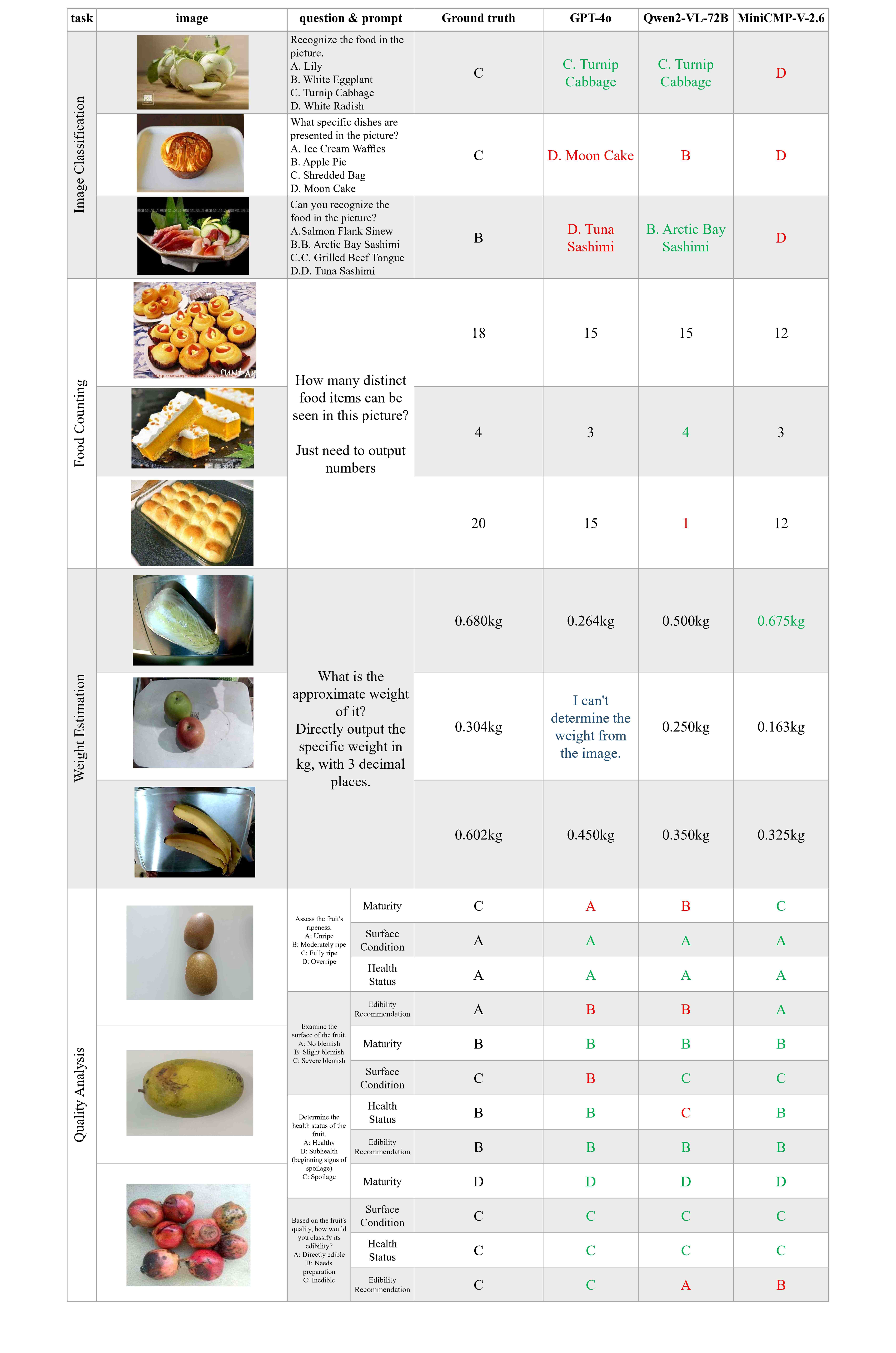}
\caption{Responses from  MiniCMP-V-2.6, Qwen2-VL-72B,GPT-4o in VQA tasks. From top to bottom, Image classification, Food counting, Weight estimation, Quality Analysis.}
\label{VQA-1}
% \vspace{-4mm}
\end{figure*}

\begin{figure*}[!t]
\centering
\includegraphics[width=1.0\linewidth]{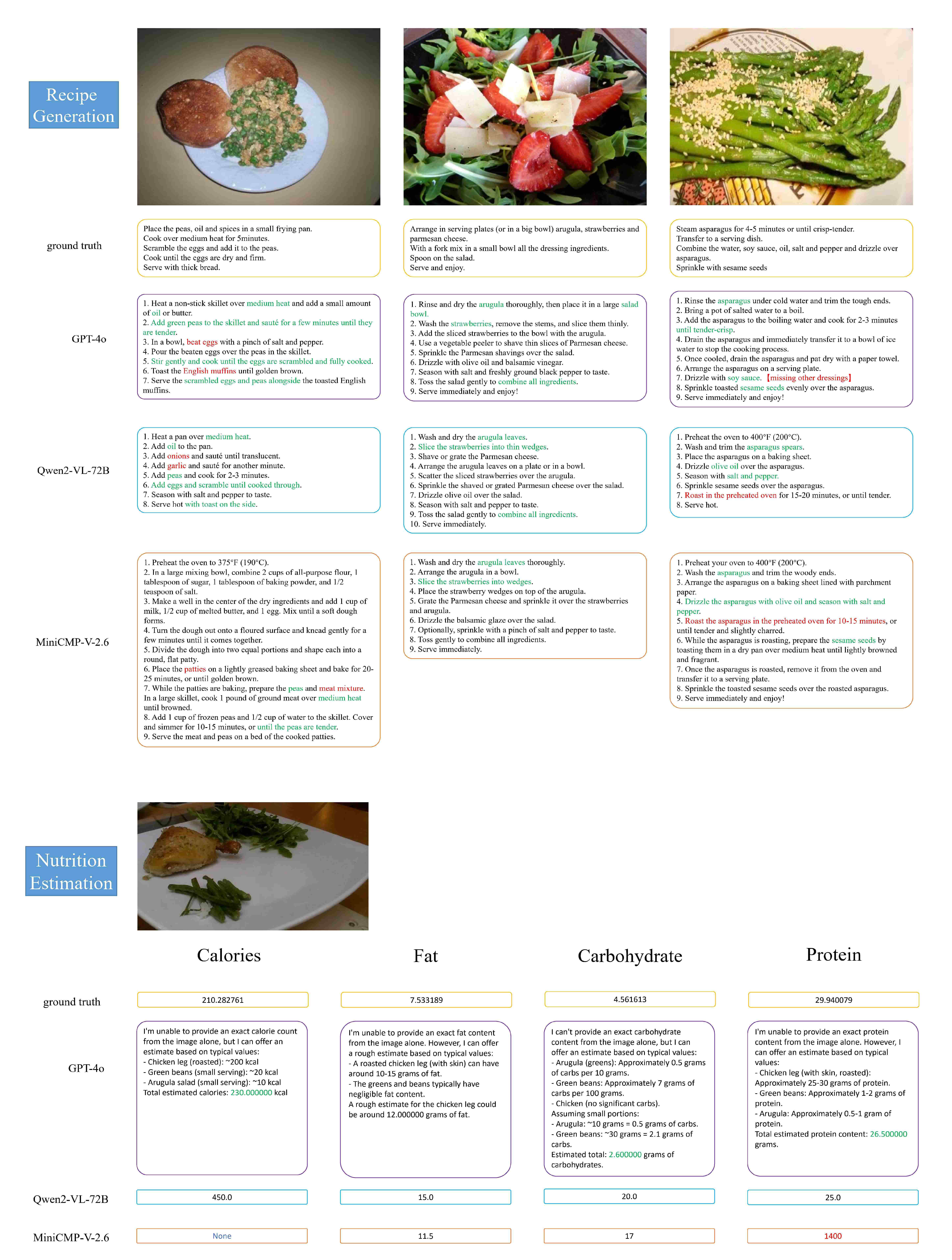}
\caption{Responses from  MiniCMP-V-2.6, Qwen2-VL-72B,GPT-4o in VQA tasks, recipe generation and nutrition estimation.}
\label{VQA-2}
% \vspace{-4mm}
\end{figure*}

\end{document}